\documentclass[journal,twoside,twocolumn]{IEEEtran}
\usepackage{amsmath,amsthm,amssymb}
\usepackage{hyperref}
\usepackage{graphicx}
\usepackage{subfig}
\usepackage{tikz}
\usepackage{algorithmicx}
\usepackage{algorithm}
\graphicspath{{./image_pdfs/}}

\begin{document} 
\title{Finding GEMS: Multi-Scale Dictionaries for High-Dimensional Graph Signals}
\author{Yael Yankelevsky,~\IEEEmembership{Student Member,~IEEE}, and Michael Elad,~\IEEEmembership{Fellow,~IEEE}% <-this % stops a space
}
\maketitle

\begin{abstract}
Modern data introduces new challenges to classic signal processing approaches, leading to a growing interest in the field of graph signal processing. 
A powerful and well established model for real world signals in various domains is sparse representation over a dictionary, combined with the ability to train the dictionary from signal examples. This model has been successfully applied to graph signals as well by integrating the underlying graph topology into the learned dictionary. 
Nonetheless, dictionary learning methods for graph signals are typically restricted to small dimensions due to the computational constraints that the dictionary learning problem entails, and due to the direct use of the graph Laplacian matrix. 
In this paper, we propose a dictionary learning algorithm that applies to a broader class of graph signals, and is capable of handling much higher dimensional data. 
We incorporate the underlying graph topology both implicitly, by forcing the learned dictionary atoms to be sparse combinations of graph-wavelet functions, and explicitly, by adding direct graph constraints to promote smoothness in both the feature and manifold domains.
The resulting atoms are thus adapted to the data of interest while adhering to the underlying graph structure and possessing a desired multi-scale property.  
Experimental results on several datasets, representing both synthetic and real network data of different nature, demonstrate the effectiveness of the proposed algorithm for graph signal processing even in high dimensions.
\end{abstract}

\section{Introduction} \label{Sec:intro}
In recent years, the field of graph signal processing has been gaining momentum. 
By merging concepts of spectral graph theory and harmonic analysis, it aims at extending classical signal processing approaches to signals having a complex and irregular underlying structure. 
Such signals emerge in numerous modern applications of diverse sources, such as transportation, energy, biological-, social-, and sensor-networks \cite{Shuman2013,Ortega2018}. 
In all these cases and many others, the underlying structure of the data could be represented using a weighted graph, such that its vertices (or nodes) represent the discrete data domain, and the edge weights reflect the pairwise similarities between these vertices. The data itself resides on the graph, that is, every graph signal is a function assigning a real value to each vertex. 

As in classical signal processing, a model for graph signals is key for handling various processing tasks, such as solving inverse problems, sampling, compression, and more. 
A popular and highly effective such model for real world signals in different domains is sparse representation \cite{Elad2010}. This model assumes the availability of a dictionary, which could be either analytic (constructed) or trained from signal examples. 
Indeed, the work reported in \cite{Zhang2012,Thanou2014,Yankelevsky2016} has deployed this breed of models to graph signals, and this paper aims at extending these contributions to allow the processing of high-dimensional graphs, which earlier methods fail to handle. 

A fundamental ingredient in the use of the sparse representations model is dictionary learning. 
Classic dictionary learning methods such as the method of optimal directions (MOD) \cite{Engan1999} and K-SVD \cite{Aharon2006} are generally structure agnostic. 
In order to better support graph signals, the work in \cite{Zhang2012,Thanou2014,Yankelevsky2016} extended these methods by integrating the underlying graph topology into the learned dictionary. 
More specifically, the work reported in \cite{Zhang2012,Thanou2014} imposed a parametric structure on the trained dictionary, relying on the graph topology. Whereas \cite{Zhang2012} learns a collection of shift-invariant graph filters, \cite{Thanou2014} restricts the dictionary to a concatenation of polynomials of the graph Laplacian matrix.

In \cite{Yankelevsky2016}, we have developed a framework for dictionary learning with graph regularity constraints in both the feature and manifold domains, which we referred to as Dual Graph Regularized Dictionary Learning (DGRDL). Furthermore, our proposed scheme suggests the additional ability of inferring the graph topology within the dictionary learning process. This is important in cases where this structure is not given, yet known to exist. 

The DGRDL algorithm and its extensions to a supervised setting \cite{Yankelevsky2017,Yankelevsky2016_ICSEE} already exhibit very good performance in various applications. 
Nevertheless, a significant limitation common to all current dictionary learning methods for graph signals (including DGRDL), is their poor scalability to high dimensional data, which is limited by the complexity of the training problem as well as by the use of the large graph Laplacian matrices. 

This limitation might be addressed by constructing analytic multi-scale transforms. 
Indeed, incorporating multi-scale properties in the dictionary is vital for representing large signals, and could reveal structural information about the signals at different resolution levels. 
Following this reasoning, classical wavelets have been generalized from the Euclidean domain to the graph setting in a number of different ways. 
Examples include the diffusion wavelets \cite{Coifman2006}, spectral graph wavelets \cite{Hammond2011}, lifting based wavelets \cite{Narang2009}, multi-scale wavelets on balanced trees \cite{Gavish2010}, permutation based wavelets \cite{Ram2012}, wavelets on graphs via deep learning \cite{Rustamov2013} and a multi-scale pyramid transform for graph signals \cite{Shuman2016}. 

Such transform-based dictionaries offer an efficient implementation that makes them less costly to apply than structure-agnostic trained dictionaries. However, while accounting for the underlying topology and possessing the desired multi-scale property, these transforms are not adapted to the given data, limiting their performance in real life applications.
\newline

In order to combine both the adaptability and the multi-scale property, while enabling treatment of higher dimensional signals, we propose infusing structure into the learned dictionary by harnessing the double sparsity framework \cite{Rubinstein2010} with a graph-Haar wavelet base dictionary. 
As such, the proposed approach benefits from the multi-scale structure and the topology-awareness that this base dictionary brings, along with the ability to adapt to the signals. It can thus be viewed as a fusion of the analytic and the trainable paradigms. 

Beyond its implicit presence through the constructed wavelet basis, the underlying data geometry is also added explicitly via direct graph regularization constraints, promoting smoothness in both the feature and manifold domains. 
Finally, we devise a complete scheme for joint learning of the graph, and hence the graph-wavelet basis, along with the dictionary. By doing so, we essentially replace the pre-constructed wavelet basis with an adaptive one, iteratively tuned along the dictionary learning process. 
The resulting algorithm, termed Graph Enhanced Multi-Scale dictionary learning (GEMS), leads to atoms that adhere to the underlying graph structure and possess a desired multi-scale property, yet they are adapted to capture the prominent features of the data of interest. 

An early version of this work appeared in \cite{Yankelevsky2018}, introducing the core idea of graph sparse-dictionary learning accompanied with preliminary experiments. This work extends the above in several important ways: (i) The introduction of the explicit regularity along with the modifications to the overall algorithm; (ii) The derivation of a joint-learning of the topology; and (iii) The addition of extensive new experiments demonstrating the strengths of the new algorithms. As these experiments show, the proposed dictionary structure brings along piecewise smoothness and localization properties, making it more suitable for modeling graph data of different nature and different dimensions. 
\newline

The outline of the paper is as follows: 
In Section~\ref{Sec:GEMS_main}, we commence by delineating the background for graph signal processing. Consequently, we revisit our DGRDL algorithm for graph signals, and present the incorporation of a sparse dictionary model, including a detailed description of the base dictionary construction procedure. 
In Section~\ref{Sec:GEMS} we consider the task of training the dictionary from examples and derive the GEMS algorithm for doing so.
Section~\ref{Sec:adaptiveL} suggests an extension that adapts the graph Laplacian, as well as the wavelet base dictionary, along the learning process. 
We then evaluate the performance of the proposed algorithm in Section~\ref{Sec:Simulation}, and conclude in Section~\ref{Sec:Conclusions}.

\section{Sparse Dictionary Learning for Graph Signals} \label{Sec:GEMS_main}

\subsection{Preliminaries} 
A weighted and undirected graph $\mathcal{G}=(\mathcal{V},\mathcal{E},W)$ consists of a finite set $\mathcal{V}$ of $N$ vertices (or nodes), a finite set $\mathcal{E}\subset \mathcal{V} \times \mathcal{V}$ of weighted edges, and a weighted adjacency matrix $W\in\mathbb{R}^{N\times N}$. The entry $w_{ij}$ represents the weight of the edge $(i,j)\in \mathcal{E}$, reflecting the similarity between the nodes $i$ and $j$. 
In general, $w_{ij}$ is non-negative, and $w_{ij}=0$ if the nodes $i$ and $j$ are not directly connected in the graph. Additionally, for undirected weighted graphs with no self-loops, $W$ is symmetric and $w_{ii}=0 \;\;\forall i$. 

The graph degree matrix $\Delta$ is the diagonal matrix whose $i$-th diagonal entry computes the sum of weights of all edges incident to the $i$-th node, i.e. having $\Delta_{ii} = \sum_j w_{ij}$. 
The combinatorial graph Laplacian matrix, representing the second-order differential operator on the graph, is then given by $L=\Delta-W$.

Given a topological graph, we refer to graph signals as functions $f:\mathcal{V}\rightarrow \mathbb{R}$ assigning a real value to each vertex. Any graph signal is therefore a vector in $\mathbb{R}^N$, whose $i$-th entry is the measurement corresponding to the $i$-th graph node.

The regularity of a graph signal $f$ can be measured using the graph Laplacian $L$ \cite{Zhou2004} in terms of the graph Dirichlet energy,
\begin{equation}
f^TLf=\frac{1}{2}\sum_{i=1}^N\sum_{j=1}^N w_{ij}(f_i-f_j)^2.
\end{equation}
When this measure of variation is small, indicating that strongly connected nodes have similar signal values, the signal is considered smooth with respect to the given graph.

\subsection{Introducing the Sparse Dictionary Model} 
The standard dictionary learning problem is formulated as
\begin{equation} \label{Eq:DictLearn}
\begin{aligned}
\arg\underset{D,X}{\min}\;&\|Y-DX\|_F^2 \\
\quad \mbox{ s.t. }\quad &\|x_i\|_0\leq T \quad \forall i, \quad\|d_j\|_2=1 \quad \forall j,
\end{aligned}
\end{equation}
where $Y\in\mathbb{R}^{N\times M}$ is the data matrix containing the training examples in its columns, $X\in\mathbb{R}^{K\times M}$ is the corresponding sparse coefficients matrix, $D\in\mathbb{R}^{N\times K}$ is an overcomplete dictionary with normalized columns (atoms), and $T$ is a sparsity threshold. The $i$-th column of the matrix $X$ is denoted $x_i$.

In order to account for the data geometry, the dual graph regularized dictionary learning (DGRDL) algorithm \cite{Yankelevsky2016} introduces graph regularity constraints in both the feature and manifold domains. 
The DGRDL problem is thus given by
\begin{equation} \label{Eq:DGRDL}
\begin{aligned}
\arg\underset{D,X}{\min}\;&\|Y-DX\|_F^2+\alpha Tr(D^TLD)+\beta Tr(XL_cX^T)\\
\quad \mbox{ s.t. }\quad &\|x_i\|_0\leq T \quad \forall i, \\
\end{aligned}
\end{equation}
where $L\in\mathbb{R}^{N\times N}$ denotes the topological graph Laplacian, accounting for the underlying inner structure of the data, and $L_c\in\mathbb{R}^{M\times M}$ is the manifold Laplacian, representing correlation between different signals within the training set. 
Imposing smoothness with respect to both graphs encourages the atoms to preserve the underlying geometry of the signals and the representations to preserve the data manifold structure. 

As mentioned in the introductory section, a significant limitation of DGRDL is poor scalability to high dimensional data, which is limited by the complexity of training, storing, and deploying the explicit dictionary $D$. 
To better accommodate higher dimensional graphs, we leverage the double sparsity approach \cite{Rubinstein2010} and propose employing a sparsity model of the dictionary atoms over a base dictionary, i.e. defining the dictionary as a product $D=\Phi A$, where $\Phi$ is some known (perhaps analytic or structured) base dictionary, and $A$ is a learned sparse matrix, having $P$ non-zeros per column.

Integrating this structure into the DGRDL scheme, we obtain the following graph-enhanced multi-scale (GEMS) dictionary learning problem:
\begin{equation} \label{Eq:GEMS}
\begin{aligned}
\arg\underset{A,X}{\min}\;&\|Y-\Phi AX\|_F^2+\alpha Tr(A^T\Phi^TL\Phi A)+\beta Tr(XL_cX^T)\\
\quad \mbox{ s.t. }\quad &\|x_i\|_0\leq T \quad \forall i, \\
\qquad \qquad &\|a_j\|_0\leq P \quad \forall j, \quad\|\Phi a_j\|_2=1 \quad \forall j.
\end{aligned}
\end{equation}
The solution can be obtained by alternating optimization over $A$ and $X$, as will be detailed in the next section.

While the double sparsity framework allows flexibility in the dimensions of $\Phi$ and $A$ and it is not generally necessary for $\Phi$ to be square, we here choose to use an orthogonal transform. Therefore, in our setting, $\Phi\in\mathbb{R}^{N\times N}$ is the base dictionary and $A\in\mathbb{R}^{N\times K}$ is a redundant ($K>N$) column-wise sparse matrix.

We emphasize that while $A$ is a redundant matrix, identical in size to the general unstructured dictionary $D$ in \eqref{Eq:DGRDL}, the dictionary update is now constrained by the number of non-zeros in the columns of $A$. Consequently, the sparse dictionary requires a training of merely $P\cdot K$ parameters rather than $N\cdot K$, where $P\ll N$. Hence learning in this case is feasible even given limited training data or high signal dimensions.

Overall, the sparse dictionary has a compact representation and provides efficient forward and adjoint operators, yet it can be effectively trained from given data even when the dimensions are very large. 
Therefore, it naturally bridges the gap between analytic dictionaries, which have efficient implementations yet lack adaptability, and standard trained dictionaries, which are fully adaptable but non-efficient and costly to deploy.

\subsection{Graph-Haar Wavelet Construction}
\label{Sec:buildPhi}

The success of the sparse dictionary model heavily depends on a proper choice of the base dictionary $\Phi$. 
In order to bring the double sparsity idea to the treatment of graph signals, we ought to define $\Phi$ such that it reflects the graph topology. 
Following our previous work \cite{Yankelevsky2018}, we choose to construct a Haar-like graph wavelet basis. 
As an initial step, and in order to expose the inherent multi-scale structure of the data, the underlying graph should be converted to a hierarchical tree by spectral partitioning. 

Spectral graph partitioning methods are commonly based on the Fiedler vector \cite{Fiedler1975}, which is the eigenvector corresponding to the smallest non-zero eigenvalue of the graph Laplacian matrix $L$. The Fiedler vector bisects the graph into two disjoint yet covering sets of nodes based on the sign of the corresponding vector entry.
Explicitly, denote the Fiedler vector for the $\ell$-th partition by $v_f^\ell$, then the bisection results in two separate sets:
\begin{equation}
\begin{aligned}
&\Omega_1^\ell = \left\lbrace i\vert v_f^\ell[i]\geq 0\right\rbrace,\\
&\Omega_2^\ell = \left\lbrace i\vert v_f^\ell[i]< 0\right\rbrace.\\
\end{aligned}
\end{equation} 
By applying the spectral bisection procedure recursively, in a coarse-to-fine manner (until reaching individual nodes or a constant-polarity Fiedler vector), full partitioning is obtained and the graph can be traversed into a hierarchical tree \cite{Simon1991}. 

We note that the Fiedler vector itself may be efficiently computed using the power-method or Lanczos algorithm \cite{Parlett1982}, without having to compute the full eigendecomposition of $L$. 
Furthermore, only a few iterations of these methods typically suffice as the bisection only depends on the sign pattern of the Fiedler vector and not on its precise values.

\begin{figure}[htb]
\begin{tikzpicture} [node distance=1cm]
\centering
\node (l0) {\includegraphics[scale=0.2,clip,trim=2.5cm 1.5cm 2cm 2cm]{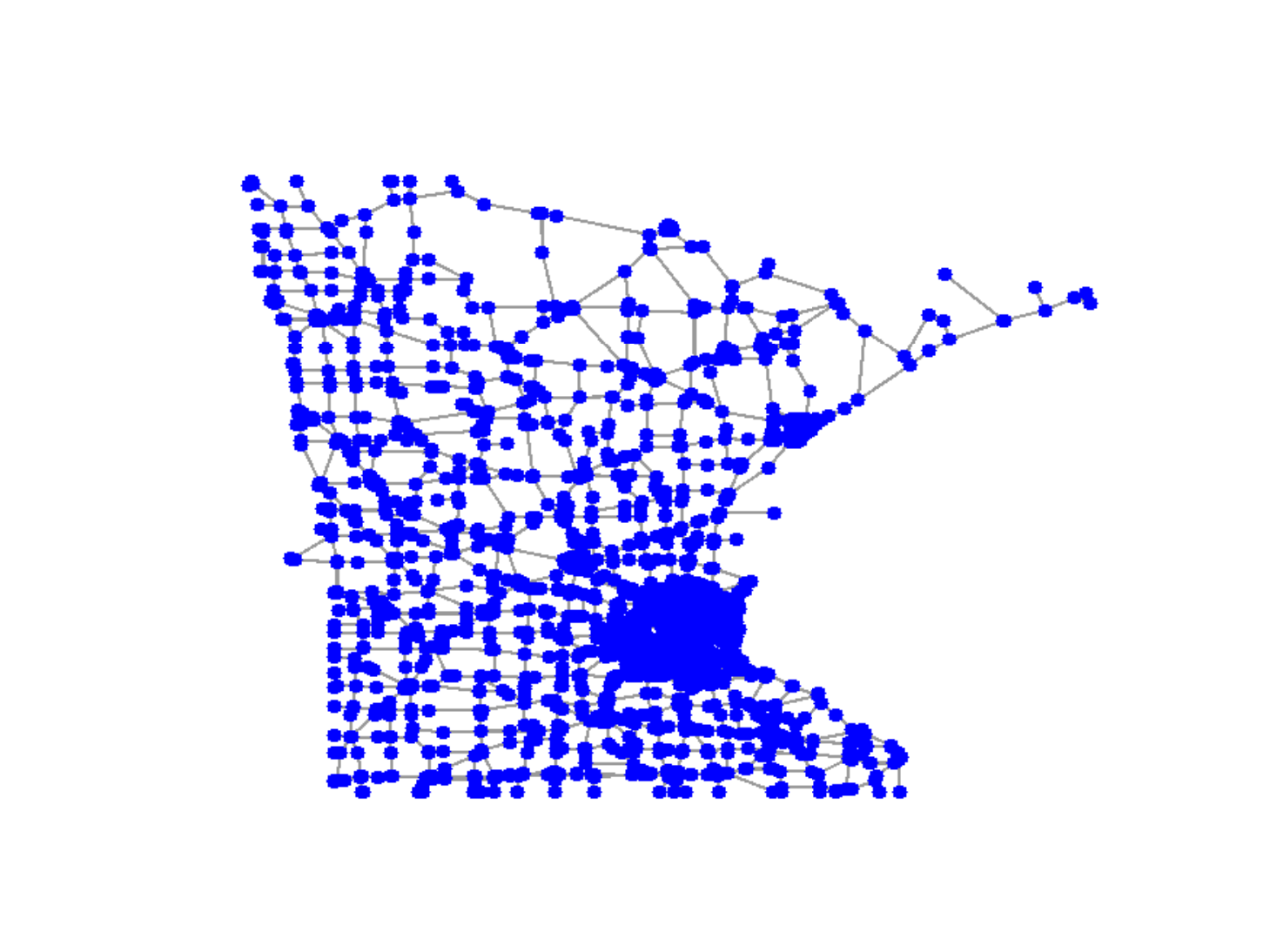}};
\node (l1a) [below of=l0, yshift=-1cm, xshift=-2cm] {\includegraphics[scale=0.2,clip,trim=2.5cm 1.5cm 2cm 3.5cm]{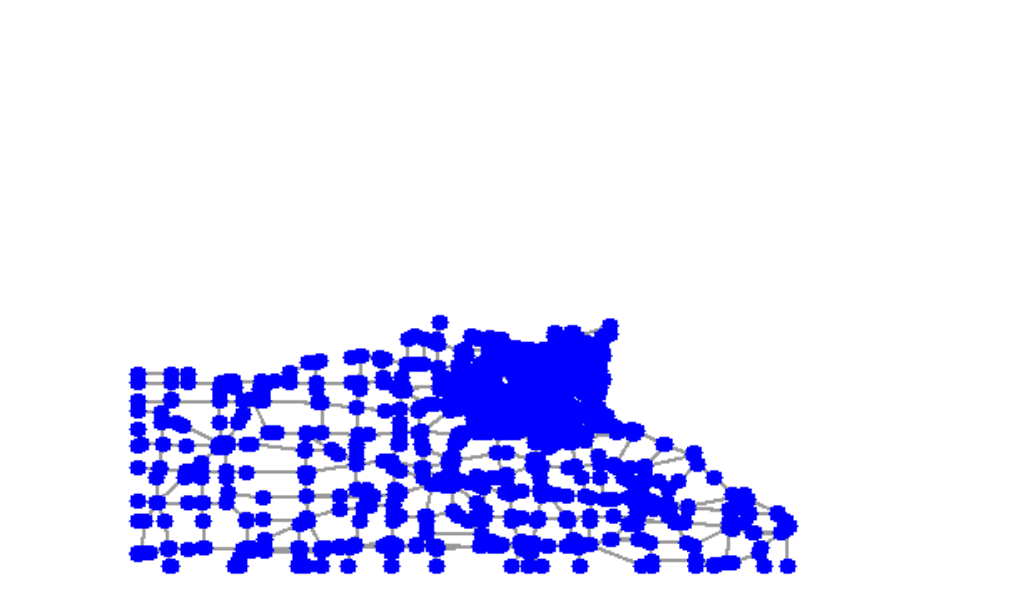}};
\node (l1b) [below of=l0, yshift=-1cm, xshift=2cm] {\includegraphics[scale=0.2,clip,trim=2.5cm 3.5cm 2cm 1.5cm]{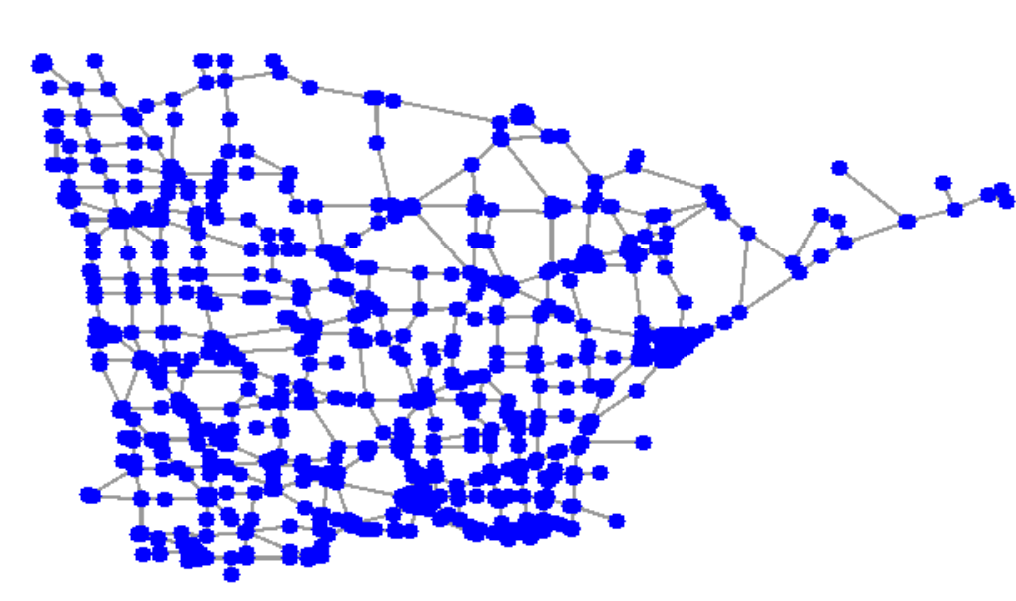}};
\node (l2a1) [below of=l1a, yshift=-1cm, xshift=-1cm] {\includegraphics[scale=0.2,clip,trim=2.5cm 1.5cm 2cm 3.5cm]{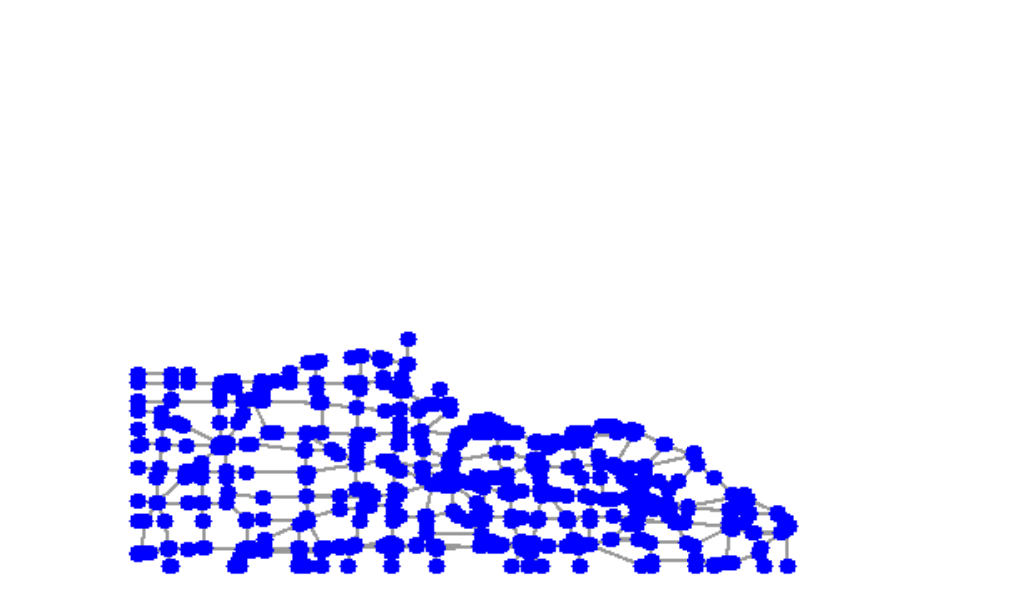}};
\node (l2b1) [below of=l1a, yshift=-1cm, xshift=1cm] {\includegraphics[scale=0.2,clip,trim=2.5cm 1.5cm 2cm 3.5cm]{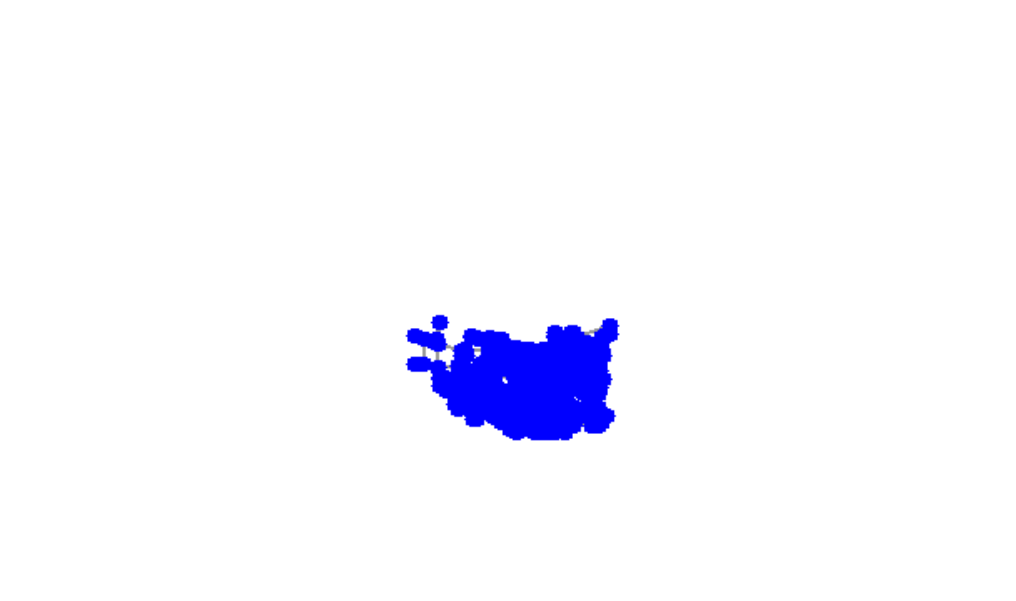}};
\node (l2a2) [below of=l1b, yshift=-1cm, xshift=-1cm] {\includegraphics[scale=0.2,clip,trim=2.5cm 3.5cm 2cm 1.5cm]{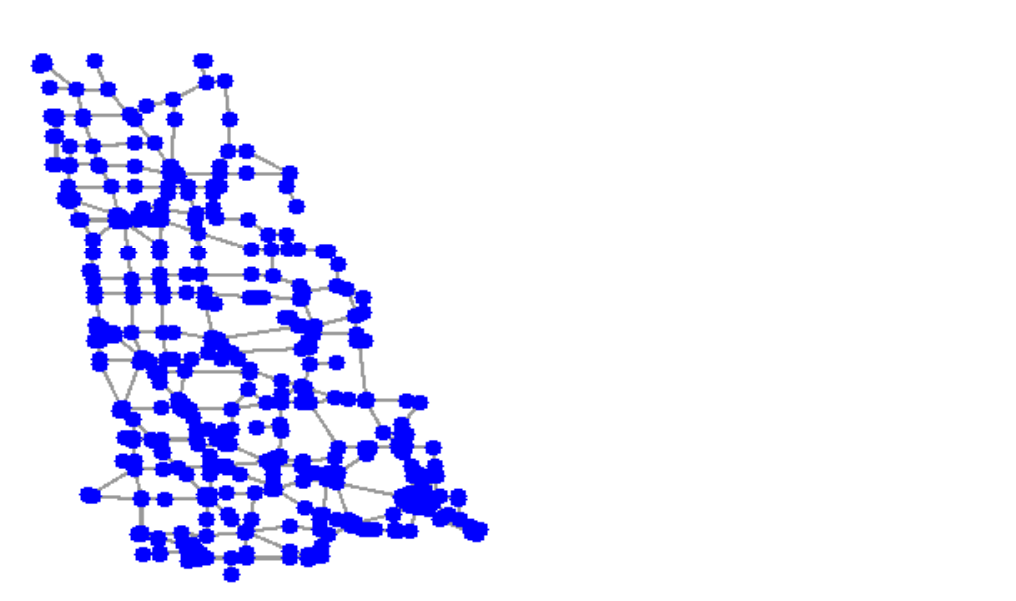}};
\node (l2b2) [below of=l1b, yshift=-1cm, xshift=1cm] {\includegraphics[scale=0.2,clip,trim=2.5cm 3.5cm 2cm 1.5cm]{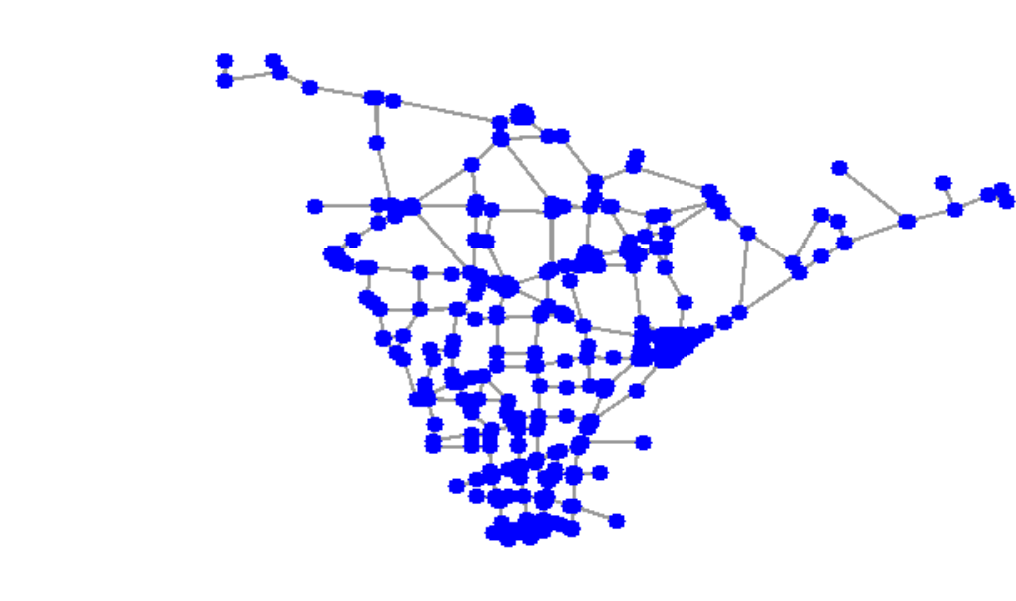}};
\draw[->] (l0) -- (l1a) node[pos=0.3,left]{\small{$\ell=1\;\;$}};
\draw[->] (l0) -- (l1b);
\draw[->] (l1a) -- (l2a1) node[pos=0.3,left]{\small{$\ell=2\;\;$}};
\draw[->] (l1a) -- (l2b1);
\draw[->] (l1b) -- (l2a2) node[pos=0.3,left]{\small{$\ell=3\;\;$}};
\draw[->] (l1b) -- (l2b2);
\end{tikzpicture}
\caption{The first partition hierarchies illustrated on the Minnesota road network graph.}
\label{Fig:Minnesota_partition}
\end{figure}
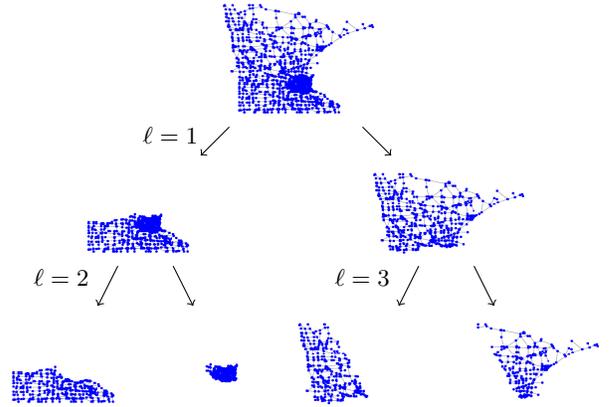

The proposed bisection approach is demonstrated in Figure~\ref{Fig:Minnesota_partition}, where the first two hierarchies of partition are presented for the Minnesota road network graph.
\newline

Equipped with the tree representation of the given data, we can now construct an orthonormal Haar-like wavelet basis in the spirit of the method proposed in \cite{Gavish2010}. 
That is, each basis function consists of constant values in each set, with the constants chosen so as to satisfy the orthogonality (meaning, in this case, that the sum of all entries should be zero) and normalization requirements. 
Explicitly, the first function is constant over the graph
\begin{equation}
\phi_0[i]=\frac{1}{\sqrt{N}}\quad \forall i,
\end{equation}
and the $\ell$-th partition induces the function 
\begin{equation}
\phi_\ell[i]=\begin{cases}
\frac{\sqrt{|\Omega_2^\ell|}}{\sqrt{|\Omega_1^\ell|}\sqrt{|\Omega_1^\ell|+|\Omega_2^\ell|}} & i\in\Omega_1^\ell,\\
-\frac{\sqrt{|\Omega_1^\ell|}}{\sqrt{|\Omega_2^\ell|}\sqrt{|\Omega_1^\ell|+|\Omega_2^\ell|}} & i\in\Omega_2^\ell,\\
0 & \mbox{else}.
\end{cases}
\end{equation}
The accumulated set of basis functions $\lbrace\phi_\ell\rbrace_{\ell}$ constitutes the columns of the matrix $\Phi$ that will serve as our base dictionary.

Not only is $\Phi$ orthogonal by construction, but also the data geometry was captured by a hierarchical tree of increasingly refined partitions. This achieves the desired localization of the constructed basis functions, and consequently, of their sparse linear combinations, which constitute the atoms of $D=\Phi A$.

\section{The Graph Enhanced Multi-Scale Dictionary Learning Algorithm (GEMS)} \label{Sec:GEMS}

\subsection{The Overall Learning Formulation}
To solve the graph-enhanced multi-scale dictionary learning problem posed in \eqref{Eq:GEMS}, we develop a K-SVD like learning scheme, based on an alternating minimization approach. 
Recall that the K-SVD iteration consists of two main steps. 
The first is sparse coding of the signals in $Y$, given the current dictionary $D=\Phi A$, to obtain $X$. 
Optimizing \eqref{Eq:GEMS} over $X$ yields the graph regularized sparse coding problem:
\begin{equation} \label{Eq:GEMS_optX}
\begin{aligned}
&\arg\underset{X}{\min}\;\|Y-\Phi AX\|_F^2+\beta Tr(XL_cX^T)\\
&\quad \mbox{ s.t. }\quad \|x_i\|_0\leq T \quad \forall i.
\end{aligned}
\end{equation}
which could be solved using our previously proposed GRSC algorithm \cite[Algorithm 2]{Yankelevsky2016} when setting $D=\Phi A$.

The second step is updating the dictionary atoms given the sparse representations in $X$. 
Note that unlike DGRDL, our structural constraint is here imposed directly on $A$, which is additionally required to preserve column-wise sparsity.
This necessitates major modifications of the dictionary update procedure.

The dictionary update is performed one atom at a time, optimizing the target function for each atom individually while keeping the remaining atoms fixed. 
To devise the update rule for the $j$-th atom, let
\begin{equation}
\begin{aligned}
\|Y-\Phi AX\|_F^2&= \|Y-\sum_{i}\Phi a_ix_i^T\|_F^2\\
&=\|E_j-\Phi a_jx_j^T\|_F^2,
\end{aligned}
\end{equation}
where $x_j^T$ denotes the $j$-th row of $X$ and we have defined the error matrix without the $j$-th atom as $E_j=Y-\sum_{i\neq j}\Phi a_ix_i^T$.

Along with the $j$-th atom we update its corresponding row of coefficients $x_j^T$. 
To preserve the representation sparsity constraints, this update uses only the subset of signals in $Y$ whose sparse representations use the current atom. 
Denote by $\Omega_j$ the indices of the subset of signals using the $j$-th atom. 

For notation simplicity, let us denote by $E,g^T,L_c^R$ the restricted versions of $E_j,x_j^T,L_c$ (respectively) limited to the subset $\Omega_j$, and let $a=a_j$. 
The target function to be minimized for updating the $j$-th atom with its corresponding coefficients is therefore
\begin{equation} 
\begin{aligned}
&\arg\underset{a,g^T}{\min}\;\|E-\Phi a g^T\|_2^2+\alpha  a^T\Phi^T L\Phi a+\beta g^T L_c^R g \\
& \qquad \mbox{ s.t. } \|a\|_0\leq P,\; \|\Phi a\|_2=1.
\end{aligned}
\end{equation}
To optimize over the atom $a$, one needs to solve
\begin{equation} \label{Eq:GEMS_optA}
\begin{aligned}
&\arg\underset{a}{\min}\;\|E-\Phi a g^T\|_2^2+\alpha  a^T\Phi^T L\Phi a \\
& \qquad \mbox{ s.t. } \;\|a\|_0\leq P,\quad \|\Phi a\|_2=1.
\end{aligned}
\end{equation}
To approximate the solution of this problem, we solve~\eqref{Eq:GEMS_optA} without the norm constraint on $\Phi a$, followed by a post-processing step that transfers energy between $a$ and $g$ to achieve $\|\Phi a\|_2=1$ while keeping the product $ag^T$ fixed.
This choice is justified if the regularization coefficient $\alpha$ is small such that the chosen support of $a$ is not impacted by the normalization.

According to Lemma 1 in \cite{Rubinstein2010}, Equation \eqref{Eq:GEMS_optA} is equivalent to 
\begin{equation} \label{Eq:GEMS_opt_atom}
\arg\underset{a}{\min}\;\|Eg-\Phi a\|_2^2+\alpha a^T\Phi^TL\Phi a \quad \mbox{ s.t. } \quad \|a\|_0\leq P
\end{equation}
as long as $g^Tg=1$. 
By applying a preprocessing step of normalizing $g$ to unit length, we can therefore further simplify the problem. This step is valid as it will only result in a scaled version of $a$, which is afterwards re-normalized anyway by balancing between $a$ and $g$.

\subsection{Dictionary Update via OMP-Like Algorithm}
One possible solution of~\eqref{Eq:GEMS_opt_atom} leverages the orthogonality of $\Phi$, by which the problem is equivalent to 
\begin{equation} 
\arg\underset{a}{\min}\;\|\Phi^TEg- a\|_2^2+\alpha a^TMa \quad\mbox{s.t.}\quad \|a\|_0\leq P
\end{equation}
where to simplify notation, we have denoted $M=\Phi^TL\Phi$.

We can devise a greedy atom pursuit algorithm for this problem, similar to the Orthogonal Matching Pursuit (OMP) \cite{Pati1993}. 
The energy to minimize for each element in the vector $a$ will here include a penalty for its correlation with all previously selected elements as reflected through the matrix $M$. 

At the $k$-th iteration, we have $\|a\|_0=k-1$ and we seek the $k$-th entry to be added.
The current residual is $r=\Phi^TEg-a$. 
The cost of choosing to add the $j$-th vector entry (assuming it was not yet included) with coefficient value $z_j$ is 
\begin{equation}
\epsilon_j = \|r-e_jz_j\|_2^2+\alpha(a+e_jz_j)^TM(a+e_jz_j),
\end{equation}
where $e_j$ denotes the $j$-th canonical vector. 
Note that the $j$-th entry in both $a$ and $r$ is assumed nulled. 

If this entry is chosen, the optimal coefficient value would be
\begin{equation}
z_j^*=\arg\underset{z_j}{\min}\;\epsilon_j=\frac{r_j-\alpha a^TM_j}{1+\alpha M_{jj}}
\end{equation}
where $M_{jj}=e_j^TMe_j$ is the $j$-th diagonal entry of $M$, and $M_j=Me_j$ is the $j$-th column of $M$.

Reorganizing $\epsilon_j$ and plugging in $z_j^*$, we obtain
\begin{equation}
\epsilon_j^* = -\frac{(r_j-\alpha a^TM_j)^2}{1+\alpha M_{jj}}+\|r\|_2^2+\alpha a^TMa.
\end{equation}
The minimum over $j$ is attained when the term $\frac{(r_j-\alpha a^TM_j)^2}{1+\alpha M_{jj}}$ is maximal\footnote{As a sanity check, notice that for $\alpha=0$ this term is simply $r_j^2$ hence the maximum is reached when $j^*=\arg\underset{j}{\max}\;|r_j|$, in consistency with the classic OMP.}, and the corresponding $j^*$-th entry will be added to the vector $a$ with entry value $z_j^*$.
Repeating the above described process for $P$ iterations, the complete sparse atom $a$ is assembled.  
\newline

The result could be further improved by adding an orthogonalization step, in which the determined support is kept fixed and the coefficient values $z_j^*$ are replaced globally using least-squares. 
Explicitly, denote by $a^R,M^R,\Psi^R$ the versions of $a,M$ and $\Psi=\Phi^TEg$ restricted to the subset of entries $\Omega$ chosen by the greedy process. Then solving
\begin{equation}
\begin{aligned}
\arg\underset{a^R}{\min}\;\|\Psi^R-a^R\|_2^2+\alpha (a^R)^TM^Ra^R
\end{aligned}
\end{equation}
leads to the optimized entries $a^R=(I+\alpha M^R)^{-1}\Psi^R$ at the support $\Omega$, composing the final atom $a$.

\subsection{Dictionary Update via ADMM}

While OMP is equipped with an efficient implementation that significantly reduces runtime, a better result can be obtained by seeking a different solution for \eqref{Eq:GEMS_opt_atom}. 
The approach we take here relies on the alternating direction method of multipliers (ADMM) \cite{Boyd2011}, and is similar in spirit to the GRSC pursuit algorithm developed for DGRDL \cite{Yankelevsky2016}. 

In this approach, we split the non-convex sparsity constraint to an auxiliary variable $b$, and Equation~\eqref{Eq:GEMS_opt_atom} is reformulated as
\begin{equation} 
\begin{aligned}
\arg\underset{a,b}{\min}\; &\|Eg-\Phi a\|_2^2+\alpha a^TMa\\
\quad\mbox{ s.t. }\quad &a=b,\;\;\|b\|_0\leq T,
\end{aligned}
\end{equation}
where we have again denoted $M=\Phi^TL\Phi$.

The augmented Lagrangian is then given by
\begin{equation} 
\mathcal{L}_{\rho}(a,b,u) = f(a)+g(b)+\rho\|a-b+u\|_2^2
\end{equation}
where $f(a)=\|Eg-\Phi a\|_2^2+\alpha a^TMa$, $g(b)=\mathcal{I}(\|b\|_0\leq P)$ for an indicator function $\mathcal{I}()$, and $u$ is the scaled dual form variable.

The iterative solution consists of sequential optimization steps over each of the variables.
Namely, in the $k$-th iteration
\begin{equation} 
\begin{cases}
a^{(k)}=\arg\underset{a}{\min}\;\|Eg-\Phi a\|_F^2+\alpha a^T M a\\
\qquad\qquad\qquad\qquad+\rho\|a-b^{(k-1)}+u^{(k-1)}\|_2^2,\\
b^{(k)}=\arg\underset{b}{\min}\;\mathcal{I}(\|b\|_0\leq P)+\rho\|a^{(k)}-b+u^{(k-1)}\|_2^2,\\
u^{(k)}=u^{(k-1)}+a^{(k)}-b^{(k)}.
\end{cases}
\end{equation}

Substituting the sub-optimization problems with their closed-form solutions results in 
\begin{equation} 
\begin{cases}
a^{(k)} = \left(\Phi^T\Phi+\alpha M +\rho I\right)^{-1}\left(\Phi^TEg+\rho(b^{(k-1)}-u^{(k-1)})\right)\\
b^{(k)}=\mathcal{S}_P\left(a^{(k)}+u^{(k-1)}\right)\\
u^{(k)}=u^{(k-1)}+a^{(k)}-b^{(k)}
\end{cases}
\end{equation}
where $\mathcal{S}_P$ is a hard-thresholding operator, keeping only the $P$ largest magnitude entries of its argument vector. 

After a few iterations, the process converges to the desired sparse atom $a=b^{(k)}$. 
Though the ADMM solution is more time consuming, it usually leads to better performance in practice compared with the greedy approach.

We should note that this algorithm comes with no convergence or optimality guarantees, since the original problem is not convex. This can be easily changed if the $\ell_0$ sparsity constraint is relaxed by an $\ell_1$ norm, thus replacing the hard-thresholding step with a soft-thresholding one.

\subsection{Updating the Coefficients}
So far, we have presented two alternative techniques for optimizing each atom of the sparse dictionary. 
Finally, having updated the atom $a$, we should update its corresponding coefficients by solving
\begin{equation} 
\arg\underset{g}{\min}\;\|E-\Phi a g^T\|_F^2+\beta g^T L_c^R g
\end{equation}
which yields
\begin{equation} 
g = \left(I+\beta L_c^R\right)^{-1}E^T\Phi a.
\end{equation}

Combining the pieces, the final atom update process consists of the following steps: 
(1) normalize $g$ to unit length; 
(2) solve \eqref{Eq:GEMS_opt_atom} using either the ADMM atom update algorithm or the OMP-like greedy pursuit proposed above; 
(3) normalize $a$ to fulfill $\|\Phi a\|_2=1$;
and (4) update $g$. 
The complete GEMS algorithm is detailed in Algorithm~\ref{Alg:GEMS}.
\begin{algorithm}[htbp]
	\caption{Graph-Enhanced Multi-Scale Dictionary Learning (GEMS)}
	\label{Alg:GEMS}
	\begin{algorithmic} [1]
		\Statex \textbf{Inputs}: signal set $Y$, base dictionary $\Phi$, initial dictionary representation  $A$, target atom sparsity $P$, target signal sparsity $T$, graph Laplacians $L$ and $L_c$
		\Statex \textbf{for} $k=1,2,...$ \textbf{do}
		\begin{itemize}
			\item \textbf{Sparse Coding:}  apply GRSC \cite{Yankelevsky2016} to solve~\eqref{Eq:GEMS_optX} for $X$
			\item \textbf{Dictionary Update:} 
			
			\textbf{for} $j=1,2,...,K$ \textbf{do}
			\begin{itemize}
				\item Identify the samples using the $j$-th atom, 
				\begin{equation} \nonumber
				\Omega_j = \left\lbrace i \;|\; 1\leq i \leq M\,,\, X_{(k)}[j,i] \neq 0 \right\rbrace
				\end{equation}
				\item Define the operator $P_j$ restricting to columns to the subset $\Omega_j$
				\item $E_j = Y-\sum_{i\neq j}{\Phi a_i x_i^T}$
				\item Set the restricted variables $E \triangleq E_jP_j$, $g^T \triangleq x_j^TP_j$ and $L_c^R \triangleq P_j^TL_cP_j$
				\item Normalize $g = \frac{g}{\|g\|_2}$
				\item Solve \eqref{Eq:GEMS_opt_atom} for $a$ (using one of the proposed methods)
				\item Normalize $a=\frac{a}{\|\Phi a\|_2}$
				\item $g=\left(I+\beta L_c^R\right)^{-1}E^T\Phi a$
				\item Plug the results $A_j = a$, $X_{(j,\Omega_j)} = g^T$
			\end{itemize}
			\textbf{end for}
		\end{itemize}
		\textbf{end for}
		\Statex \textbf{Outputs:} $A,X$
	\end{algorithmic}
\end{algorithm}

\section{Adaptive Base Dictionary} \label{Sec:adaptiveL}
In cases where the true underlying graph is unknown, it can be constructed or inferred from the data. 
Several attempts have recently been made to learn the underlying graph from data observations \cite{Pavez2016,Dong2016,Kalofolias2016,Segarra2017,Egilmez2017}. 
In this work, similarly to the approach we proposed in \cite{Yankelevsky2016}, we could leverage the trained dictionary, that already processed the input and captured its essence, to adapt and improve the graph Laplacian estimation. That is, the graph is learned jointly with the dictionary rather than being learned directly from the observed signals.

The extension of the proposed GEMS algorithm for this case is straightforward. 
When the graph Laplacian is unknown, we initialize it from the training data $Y$ using some common construction (such as a Gaussian kernel). 
Based on this initial $L$, we construct the base dictionary $\Phi$ as described in Section~\ref{Sec:buildPhi} and run a few iterations of the GEMS algorithm, without reaching full convergence. Having at hand an updated sparse matrix $A$, and therefore an updated effective dictionary $D$, we could optimize the graph Laplacian $L$ such that it leads to smoother atoms over the graph. Adding some requirements to normalize $L$ and make it a valid graph Laplacian matrix, the resulting optimization problem is
\begin{equation} \label{Eq:GEMS_optL}
\begin{aligned}
&\arg\underset{L}{\min}\;\alpha Tr(A^T\Phi^TL\Phi A)+\mu\|L\|_F^2\\
&\mbox{ s.t. }\quad L_{ij}=L_{ji}\leq 0 \; (i\neq j), \;L\underline{1}=\underline{0}, \;Tr(L)= N.
\end{aligned}
\end{equation}
Note that this is in fact the same problem defined in \cite{Yankelevsky2016} for the setting $D=\Phi A$.

By vectorizing $L$, Equation~\eqref{Eq:GEMS_optL} can be cast as a quadratic optimization problem with linear constraints, which could be solved using existing convex optimization tools. As the computational complexity scales quadratically with the number of nodes $N$, for very large graphs an approximate solution may be sought based on splitting methods or using iterative approaches.

An important consequence of the graph optimization is that given an updated $L$, the base dictionary $\Phi$ could now be refined as well. 
By doing so, we effectively replace the fixed graph-wavelet basis with an adaptive one, which is iteratively tuned along with the dictionary learning process, thus adding yet another level of flexibility to the proposed scheme. 
Having reconstructed $\Phi$, the GEMS algorithm can be resumed for several more iterations. This process of updating $L$, refining $\Phi$ and applying GEMS can be repeated until converging to a desired output.

It should be emphasized that the Laplacian optimization may be applied to the manifold Laplacian $L_c$ as well in a similar manner. 
\newline

Before diving into the experimental section, we briefly discuss a special setting of the proposed GEMS algorithm obtained by omitting the explicit regularizations, i.e. setting $\alpha=\beta=0$.
This choice alleviates the additional complexity of updating the atoms and so further improves the scalability of this method and enables treatment of very large graphs. 
The optimization problem for this setting reduces to
\begin{equation} 
\begin{aligned}
\arg\underset{A,X}{\min}\;&\|Y-\Phi AX\|_F^2\\
\quad \mbox{ s.t. }\quad &\|x_i\|_0\leq T \quad \forall i, \\
\qquad \qquad &\|a_j\|_0\leq P \quad \forall j, \quad\|\Phi a_j\|_2=1 \quad \forall j.
\end{aligned}
\end{equation}
Nevertheless, the graph Laplacian $L$ is still accounted for implicitly through the construction of $\Phi$. 
Therefore, an optimized Laplacian could still enhance our method even in this setting: by gradually refining the base dictionary along the training process, this special setting instigates an adaptive graph-Haar wavelet dictionary.
In that sense, this configuration can be seen as an adaptive version of SDL \cite{Yankelevsky2018}, in which the base dictionary $\Phi$ is updated along the training process.  
We shall henceforth refer to this very high-dimensional setting as GEMS-HD.

\section{Experiments and Applications} \label{Sec:Simulation}
In this section, we demonstrate the effectiveness of the proposed GEMS algorithm on synthetic examples of piecewise-smooth nature and on real network data, and show its potential use in various data processing and analysis applications.

\subsection{Synthetic Experiment}
We first carry out synthetic experiments, similar to the ones described in \cite{Yankelevsky2016,Yankelevsky2018}. 
However, to corroborate the applicability of GEMS to a broader class of graph signals, the generated data here complies with a piecewise-smooth model rather than the global-smooth one used in \cite{Yankelevsky2016,Yankelevsky2018}.

Initially, we generated a random graph consisting of $N$ randomly distributed nodes. 
The edge weights between each pair of nodes were determined based on the Euclidean distances between them $d(i,j)$ and using the Gaussian Radial Basis Function (RBF) $w_{ij}=\exp\left(\frac{-d^2(i,j)}{2\sigma^2}\right)$ with $\sigma=0.5$. 

For the data generation, we started by simulating two sets of globally-smooth graph signals. 
Each such set was created by randomly drawing an initial matrix $Y_0\in\mathbb{R}^{N\times 10N}$ and solving 
\begin{equation}
\arg\underset{Y}{\min}\; \|Y-Y_0\|_F^2+\lambda Tr(Y^TLY),
\end{equation}
which yields smoothed signals $Y=\left(I+\lambda L\right)^{-1}Y_0$.

Given two such data matrices $Y_1$ and $Y_2$, we combined them to generate piecewise-smooth graph signals. For that purpose, a local neighborhood was randomly chosen for each signal, and its measurements in that region were taken from $Y_2$ while the rest were taken from $Y_1$. 
Consequently, each signal was normalized to have unit norm. 
A subset of $40\%$ of the generated signals was used for training, leaving the rest for testing.

Using this training data, several dictionaries were learned including the K-SVD \cite{Aharon2006}, the graph Polynomial dictionary \cite{Thanou2014}, DGRDL \cite{Yankelevsky2016}, and the proposed GEMS with a graph-Haar base dictionary constructed as in Section~\ref{Sec:buildPhi}. 
Additionally, we trained the very high-dimensional mode GEMS-HD, for setting $\alpha=\beta=0$. 
For a fair comparison, all these dictionaries are of the same size, $N\times 2N$. We also evaluated a direct use of the constructed graph wavelet basis $\Phi$, whose size is $N\times N$.

Two setups were tested: the first with a moderate size graph of $N=256$ nodes, and the second with a high-dimensional graph containing $N=4096$ nodes.
The dictionaries were trained with a fixed number of non-zeros in the sparse coding stage ($T=12$ and $T=25$, respectively). For the presented variants of GEMS, the respective sparsity levels of the dictionary $A$ were set to $P=12$ for the medium graph and $P=40$ for the large one.
\newline

The dictionaries were first compared by their ability to obtain the best m-term approximation of the test data (for different sparsity levels, both smaller and larger than the number of non-zeros used during training), and performance was measured in terms of the normalized Root Mean Squared Error (RMSE), $\frac{1}{\sqrt{NM}}\|Y-DX\|_F$. 

\begin{figure}[htb] 
	\centering
	\subfloat[Representation error for $N=256$]{
		\centering \includegraphics[scale=0.38,clip,trim=0.8cm 0.3cm 0.8cm 1.03cm]{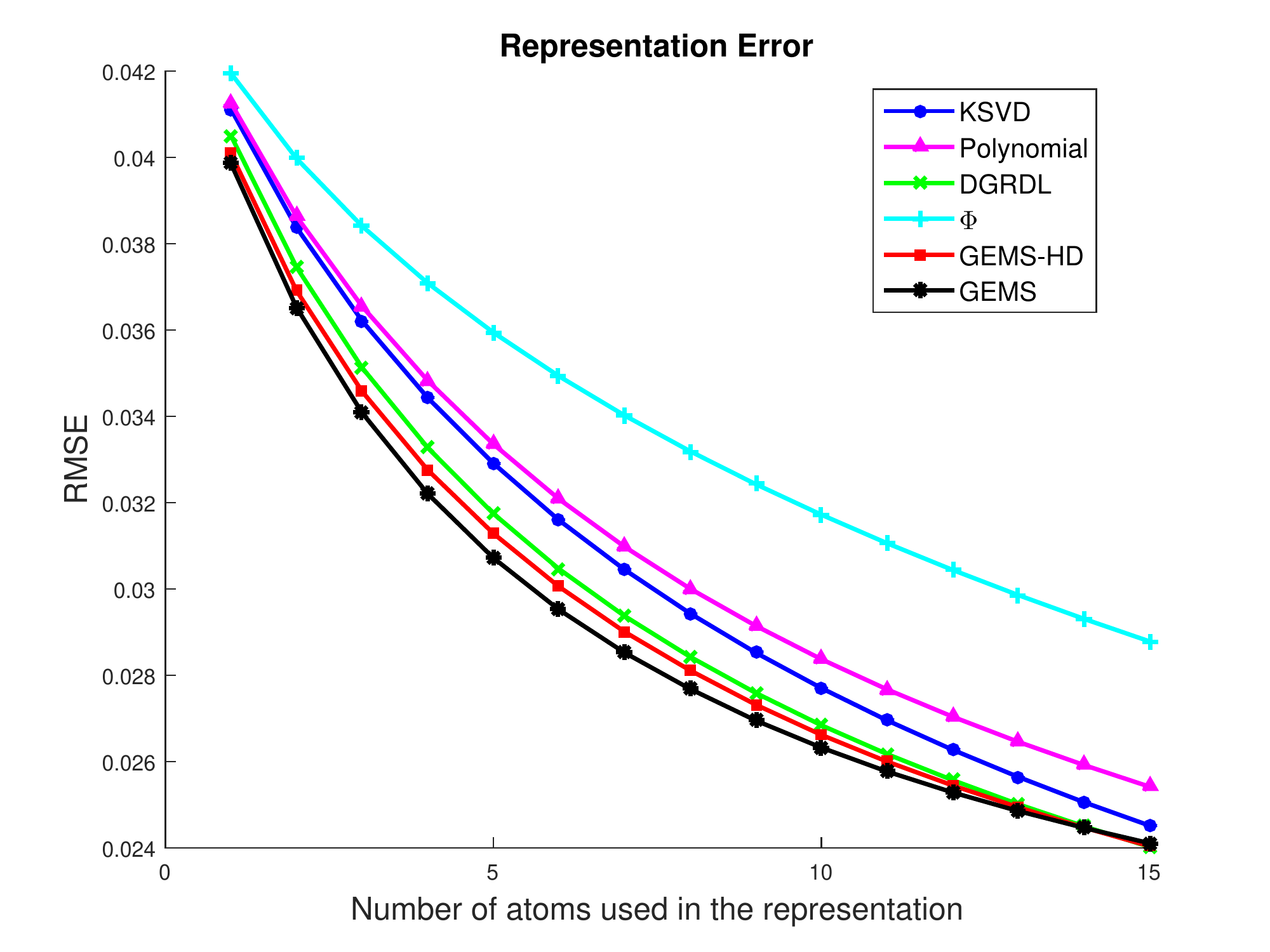}		
		\label{Fig:synthetic_rep}
	}\\
	\subfloat[Representation error for $N=4096$]{
		\centering \includegraphics[scale=0.38,clip,trim=0.8cm 0.3cm 0.8cm 1.03cm]{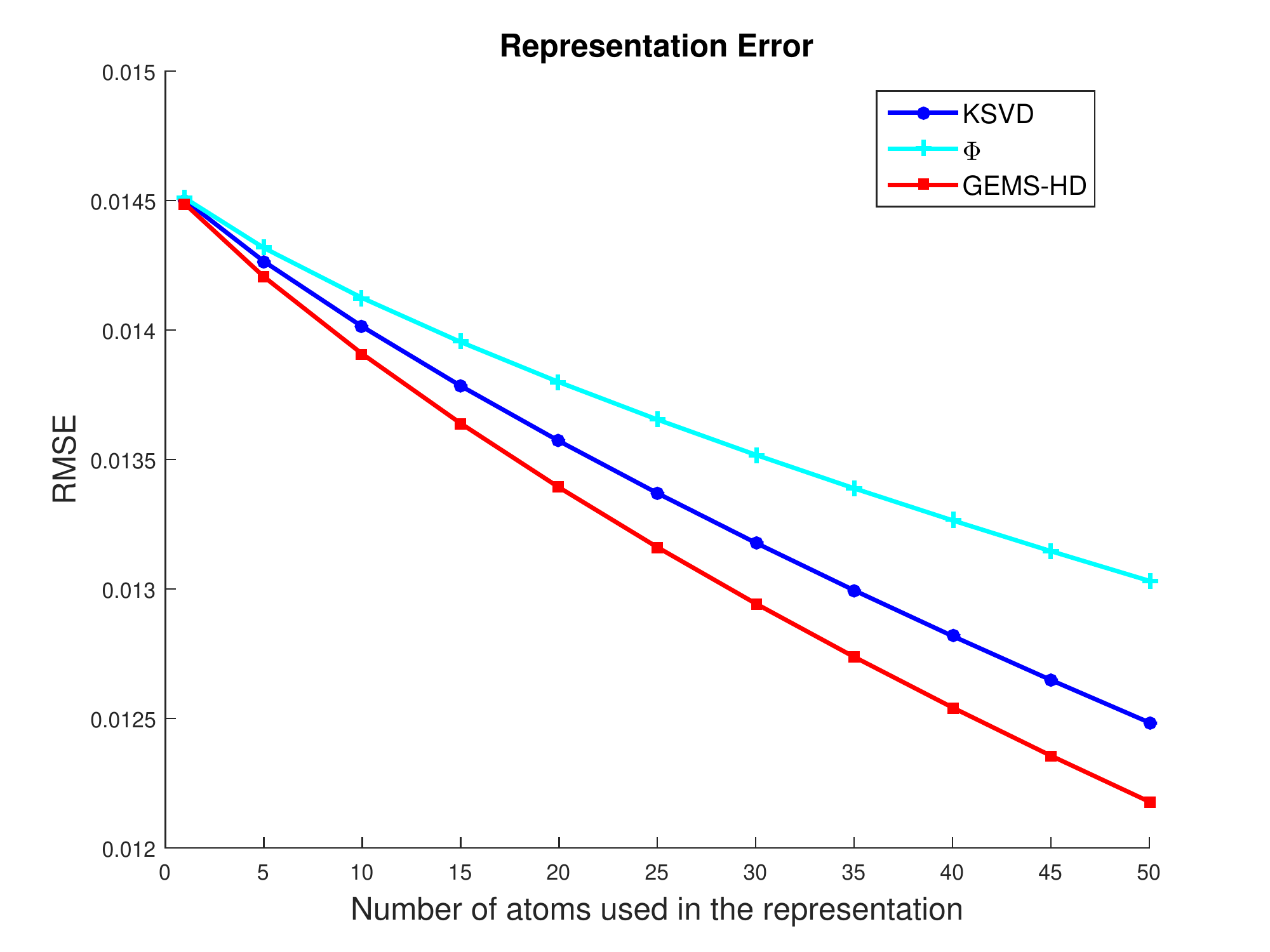}		
		\label{Fig:synthetic_rep_large}
	}
	\caption[]{Comparison of the learned dictionaries in terms of normalized RMSE for representing synthetic data of different dimensions with various sparsity levels.}
	\label{Fig:synthetic_exp_results_rep}
\end{figure}

The representation errors presented in Figure~\ref{Fig:synthetic_rep} show that for a moderate size graph, the proposed GEMS yields lower errors compared with K-SVD, DGRDL, the Polynomial method and the graph-Haar wavelet basis $\Phi$, for all evaluated sparsity levels. 
Furthermore, the complete GEMS scheme offers an additional improvement over GEMS-HD, that only accounts for the graph implicitly.

The representation errors obtained for a large graph setting are presented in Figure~\ref{Fig:synthetic_rep_large}.
For this data dimension, the Polynomial dictionary and DGRDL can no longer train in reasonable time, and were therefore omitted from the comparison. For computational reasons, GEMS was also trained only in the GEMS-HD mode. 
Nevertheless, it still outperforms K-SVD and the graph wavelet base dictionary $\Phi$, demonstrating the scalability of the proposed method to high dimensional data.

Next, the performance of the trained dictionaries was evaluated for the common task of signal denoising, by adding Gaussian noise of different levels $\sigma_n$ to the test signals and comparing recovery using each of the dictionaries in terms of the normalized RMSE. Assuming a noisy test signal is modeled as $y_i = Dx_i + n_i$ where $n_i$ denotes the added noise, its denoised version $\hat{y}_i = D\hat{x}_i$ is obtained by seeking the sparse approximation of $y_i$ (denoted $\hat{x}_i$) over each dictionary $D$ with a known sparsity level $T$.

\begin{figure}[htb] 
\centering
	\subfloat[Denoising error for $N=256$]{
		\centering \includegraphics[scale=0.38,clip,trim=0.8cm 0.3cm 0.8cm 1.03cm]{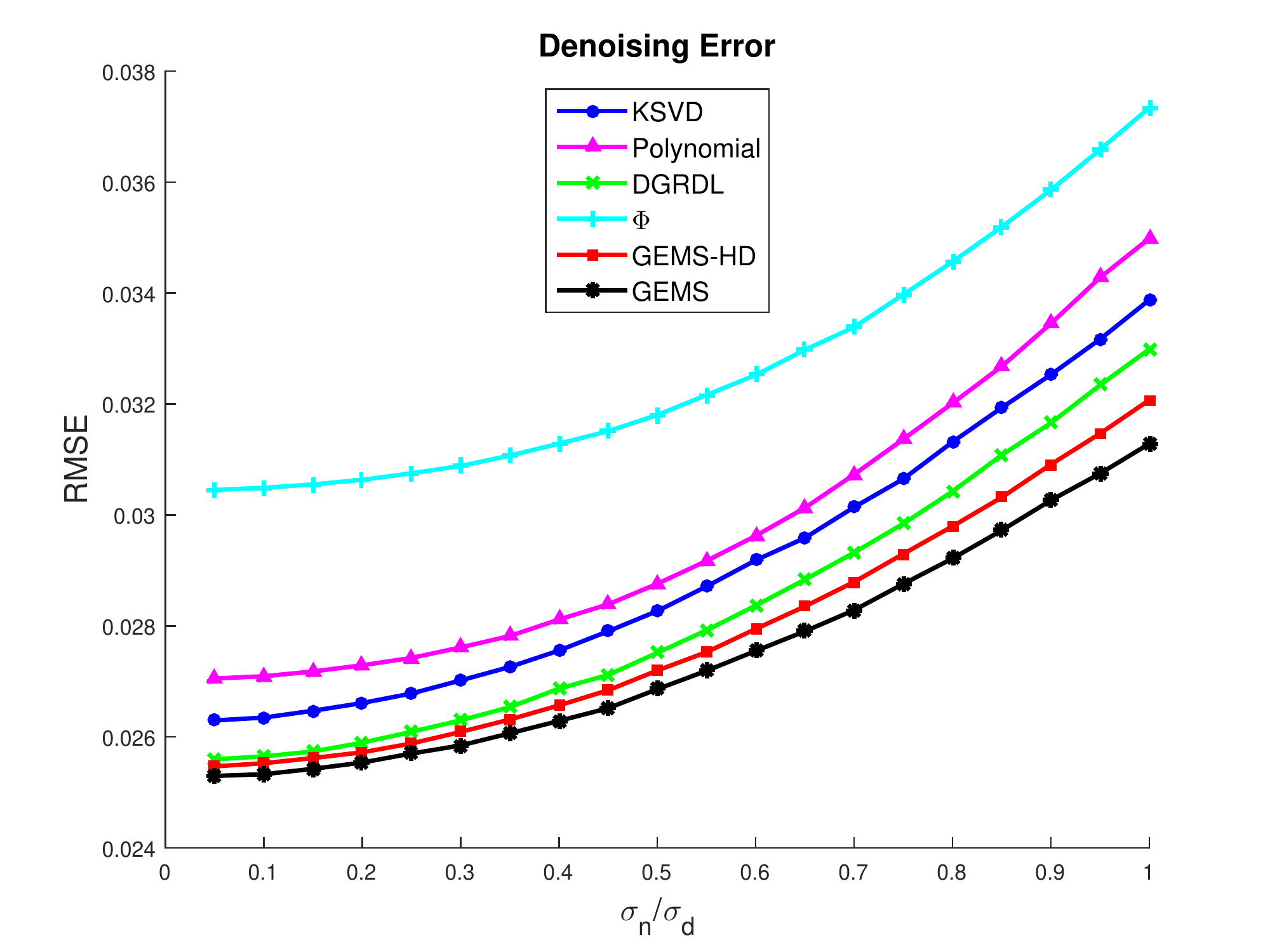}
		\label{Fig:synthetic_den}
	}\\
	\subfloat[Denoising error for $N=4096$]{
		\centering \includegraphics[scale=0.38,clip,trim=0.8cm 0.3cm 0.8cm 1.03cm]{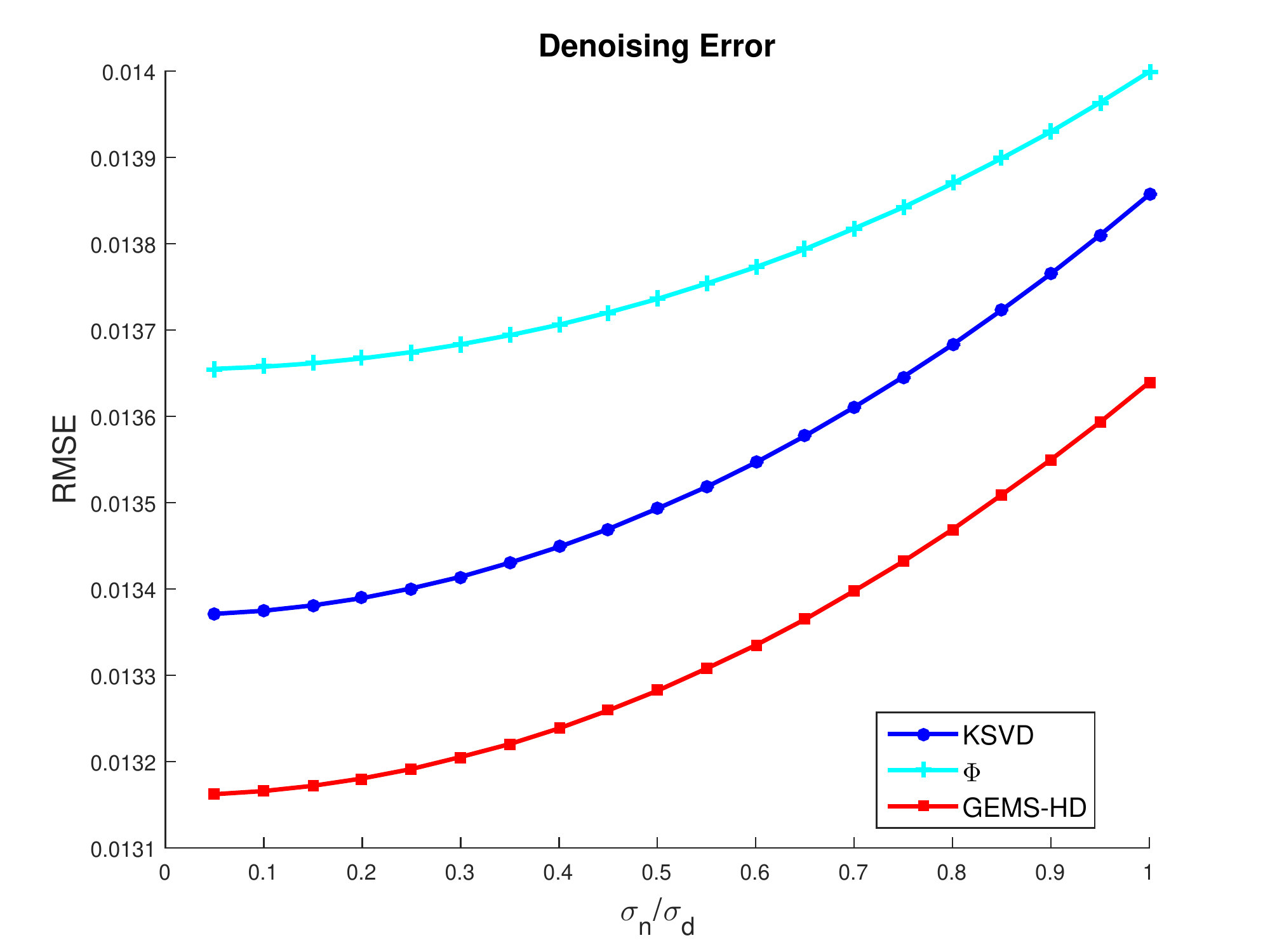}	
		\label{Fig:synthetic_den_large}
	}
	\caption[]{Comparison of the learned dictionaries in terms of normalized RMSE for the task of synthetic data denoising with different noise levels $\sigma_n$ with respect to the data standard deviation $\sigma_d$.}
	\label{Fig:synthetic_exp_results_den}
\end{figure}

The results of this experiment are depicted in Figure~\ref{Fig:synthetic_exp_results_den}. Similarly to the previous experiment, these results show that GEMS outperforms the other dictionary models for all the tested noise levels, and offers a performance boost for graph signal denoising even in high dimensions. 
\newline

\begin{figure*}[htbp]
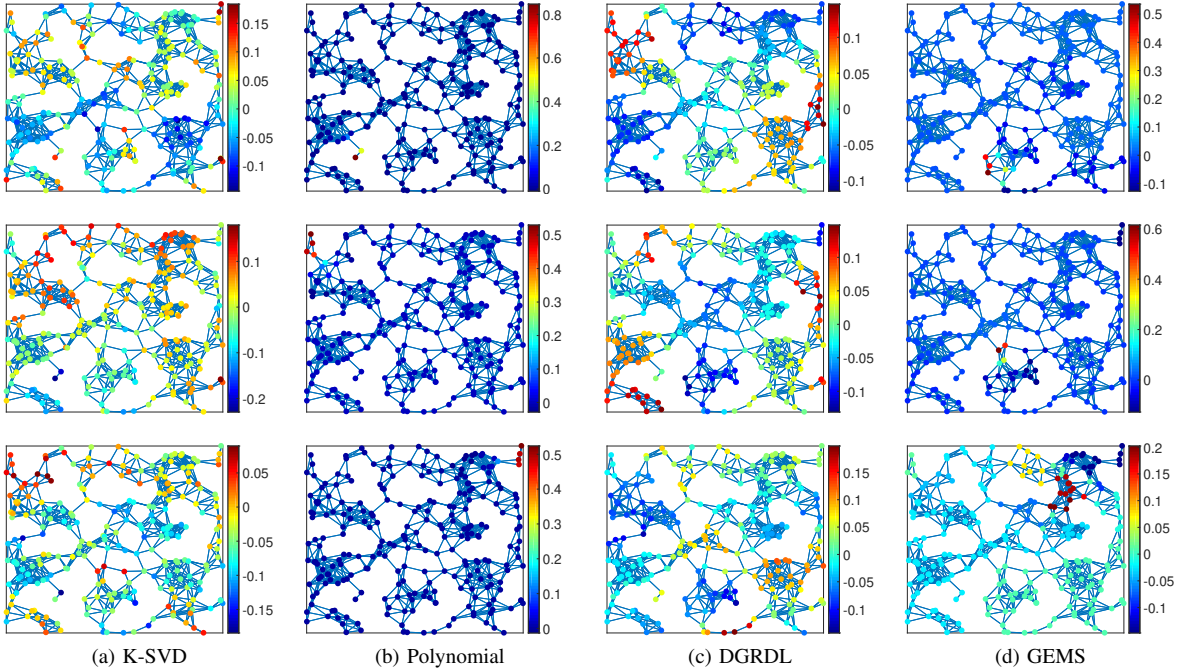

\centering
\foreach \dict in {KSVD,Poly,DGRDL,GEMS} {
	\centering \subfloat{
		\centering \includegraphics[scale=0.2,clip,trim=2.2cm 1.5cm 0cm 0.8cm]{\dict_Top_Atoms_1}
	}
}\\
\foreach \dict in {KSVD,Poly,DGRDL,GEMS} {
	\centering \subfloat{
		\centering \includegraphics[scale=0.2,clip,trim=2.2cm 1.5cm 0cm 0.8cm]{\dict_Top_Atoms_2}
	}
}\\
\setcounter{subfigure}{0}
\foreach \dictname/\dict in {K-SVD/KSVD,Polynomial/Poly,DGRDL/DGRDL,GEMS/GEMS} {
	\centering \subfloat[][\dictname]{
		\centering \includegraphics[scale=0.2,clip,trim=2.2cm 1.5cm 0cm 0.8cm]{\dict_Top_Atoms_3}
	}
}
\caption[]{Demonstrating the top 3 atoms used of each evaluated dictionary. Each column refers to a different dictionary (from left to right): K-SVD \cite{Aharon2006}, Polynomial \cite{Thanou2014}, DGRDL \cite{Yankelevsky2016} and GEMS. It can be observed that GEMS yields atoms that obey a piecewise-smooth model as desired.}
\label{Fig:compareAtomsSynthetic}
\end{figure*}

Additionally, we verify that the proposed dictionary indeed results in more localized atoms by visualizing the top 3 used atoms of each of the trained dictionaries. As can be observed in Figure~\ref{Fig:compareAtomsSynthetic}, although all dictionaries were trained from piecewise-smooth graph signals, the atoms learned by K-SVD are unstructured and possess a random appearance, the Polynomial atoms are extremely sparse and localized, and the DGRDL atoms vary more gradually than their K-SVD counterparts, yet they span the support of the entire graph. The atoms learned by GEMS are more localized and structured compared with those learned by K-SVD and DGRDL, though not as localized as the Polynomial atoms. GEMS thus offers a balance between the localization and smoothness properties, yielding atoms that have the desired piecewise-smooth nature. 

\subsection{Flickr Data}
In the sequel, the proposed method was evaluated on real network data from the Flickr dataset. 
The dataset consists of 913 signals, representing the daily number of distinct Flickr users that have taken photos at different geographical locations around Trafalgar Square in London, between January 2010 and June 2012. 
An area of approximately $6\times 6$ km was covered by a grid of size $16\times 16$, to a total of $N=256$ nodes. 
The initial graph Laplacian $L$ was designed by connecting each node and its 8 nearest neighbors, setting the edge weights to be inversely proportional to the Euclidean distance between the nodes. 

\begin{figure*}[htbp]
	\centering
	\foreach \sig in {159,584,758} {			
		\centering \subfloat{
			\centering \includegraphics[scale=0.3,clip,trim=1.2cm 0.8cm 1.0cm 0cm]{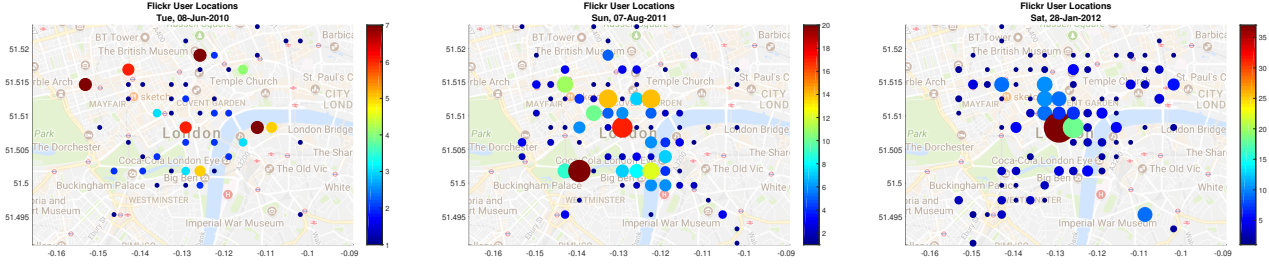}
		}
	}	
	\caption[]{Characteristic graph signals demonstrating the daily number of distinct Flickr users that have taken photos at different locations in London. The size and color of each circle indicate the signal value at that graph node.}
	\label{Fig:FlickrSignals}
\end{figure*}

Each photo acquisition was allocated to its nearest grid point, so that the graph signals represent the spatially aggregated daily number of users taking photos near each grid location. 
A random subset of 700 signals constitutes the training set, and the rest were used for testing. 
All signals were normalized with respect to the one having the maximal energy. 
Some typical signals from the Flickr dataset are illustrated in Figure~\ref{Fig:FlickrSignals}.

For this dataset, the proposed GEMS dictionary was again compared with K-SVD \cite{Aharon2006}, the graph Polynomial dictionary \cite{Thanou2014} and DGRDL \cite{Yankelevsky2016}, as well as with the constructed graph-Haar wavelet basis $\Phi$. 
All evaluated dictionaries are of the same size of $N\times 2N$ (with the exception of the orthogonal basis $\Phi$ whose dimensions are $N\times N$) and sparsity thresholds of $T=3$ and $P=10$ were used for training.
\newline

Similarly to the synthetic experiment, the different dictionaries were evaluated on two tasks: their ability to represent the test set data with different sparsity levels (number of used atoms), and their performance in signal denoising with different noise levels.

\begin{figure}[htb] 
	\centering
	\subfloat[]{
		\includegraphics[scale=0.38,clip,trim=0.8cm 0.3cm 0.8cm 0.67cm]{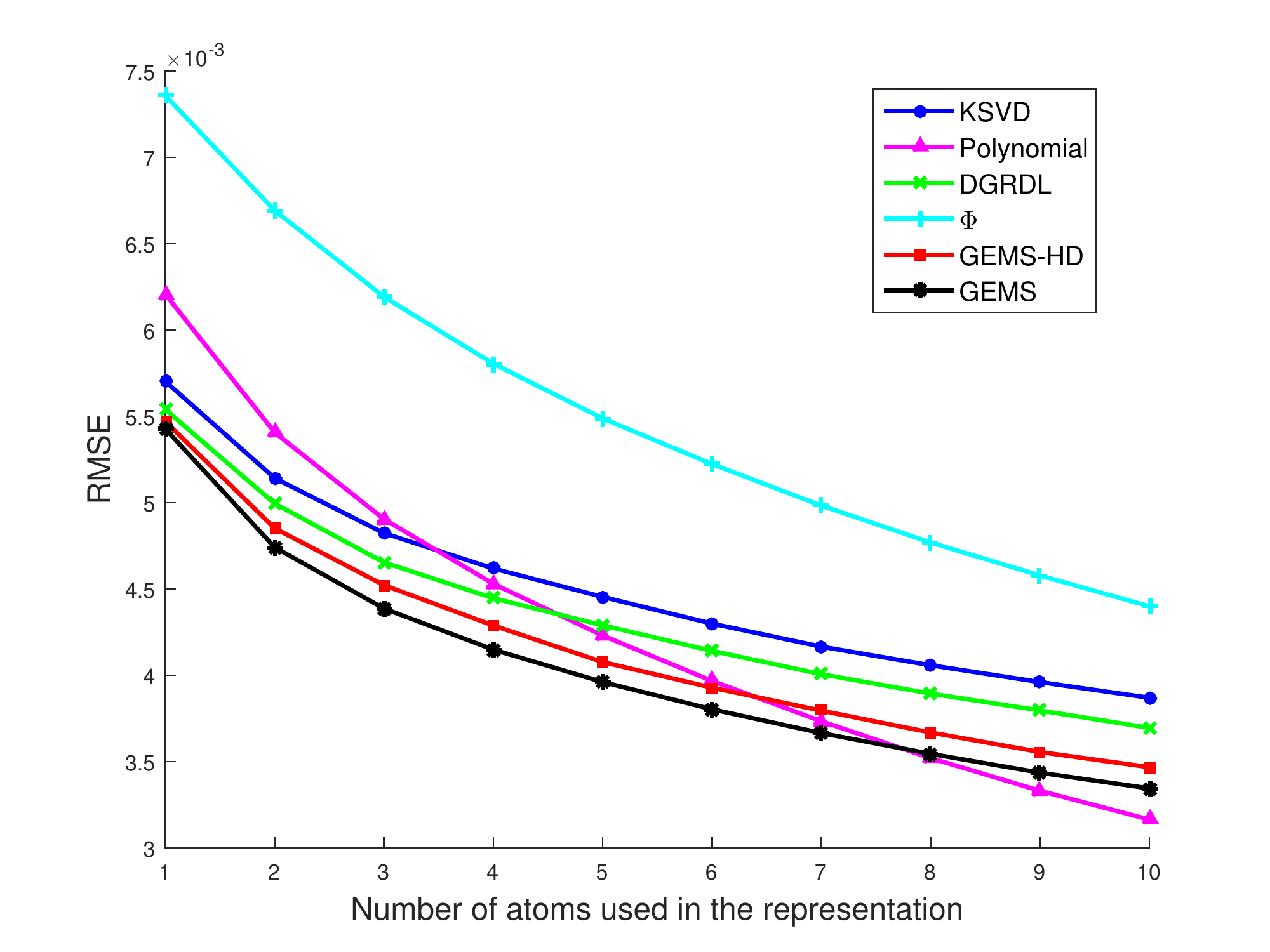}		
		\label{Fig:flickr_rep}
	}\\
	\subfloat[]{
		\includegraphics[scale=0.38,clip,trim=0.8cm 0.3cm 0.8cm 0.67cm]{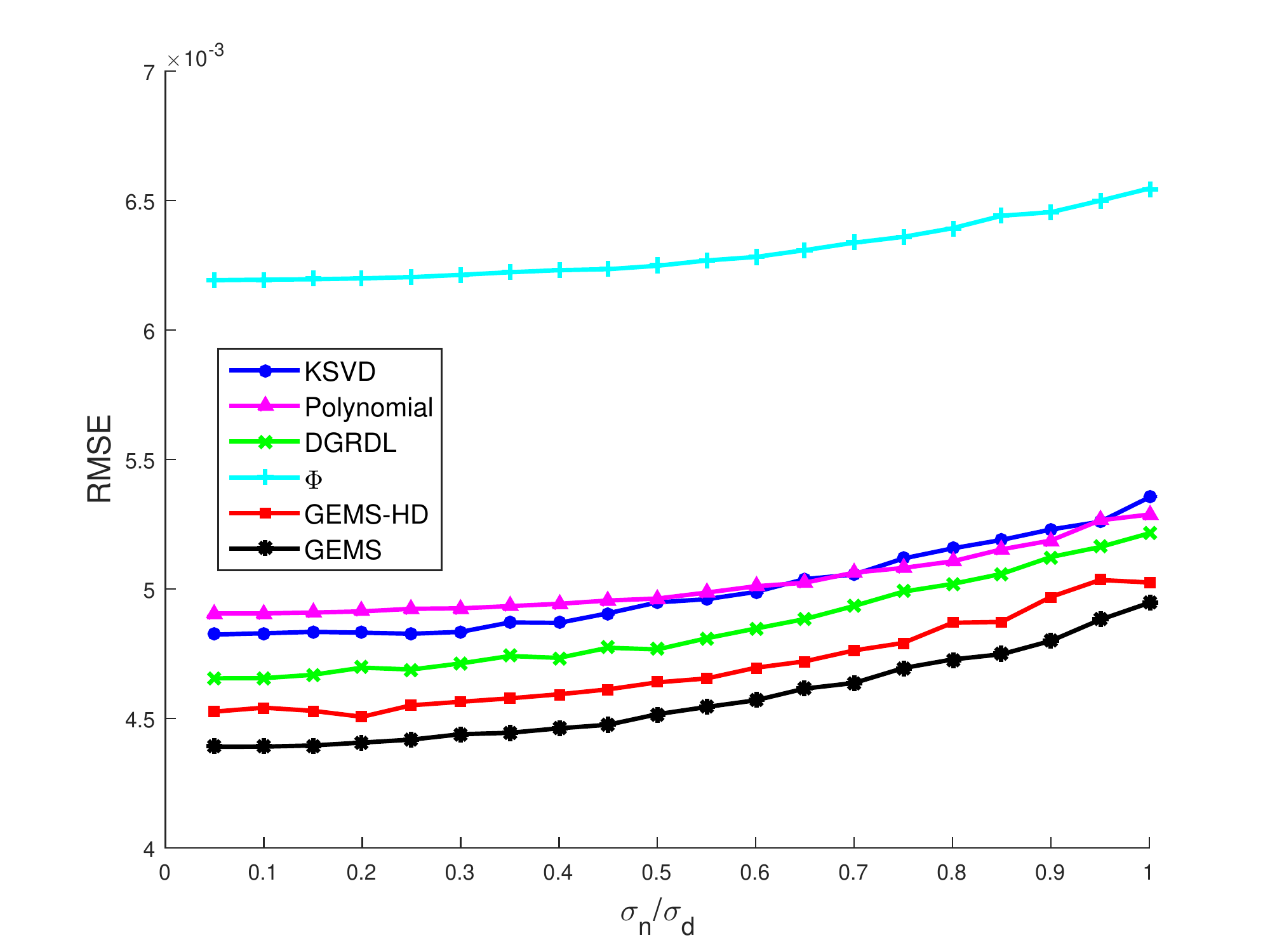}				
		\label{Fig:flickr_den}
	}
	\caption[]{Comparison of the learned dictionaries in terms of normalized RMSE for different applications tested on the Flickr dataset: \subref{Fig:flickr_rep} representation error for different sparsity levels, \subref{Fig:flickr_den} denoising error for different noise levels $\sigma_n$ with respect to the data standard deviation $\sigma_d$.}
	\label{Fig:flickr_rep_den}
\end{figure}

The representation errors for this dataset are presented in Figure~\ref{Fig:flickr_rep}, and the corresponding denoising errors in Figure~\ref{Fig:flickr_den}.

It can be observed that in both tasks, for all sparsity levels and all noise levels tested, GEMS yields significantly lower errors compared with K-SVD, the Polynomial graph dictionary, and DGRDL. These results coincide with those obtained for the synthetic experiment. 
The only exception is the approximation error using a larger number of atoms ($T>8$), for which the Polynomial dictionary achieves slightly better results than GEMS. Recall, however, that the Polynomial dictionary training is much more complex and its runtime is substantially longer, making its use impractical for larger dimensions. 

It should also be emphasized that the performance of GEMS is expected to further improve as the training set becomes scarce. 

Moreover, the results could be improved by re-training the dictionaries for every sparsity level. Instead, training was performed once for a fixed $T$ and the generalization ability of the dictionaries was challenged by evaluating them using different (both smaller and larger) sparsity levels. Nevertheless, as the experimental results demonstrate, the trained GEMS model fits the data very well even in this setting.

\subsection{Uber Pickups in New York City}
Next, we consider a larger real network dataset of Uber pickups in New York City \cite{uberDB}. 
This dataset contains information on over 4.5 million Uber pickups in New York City from April to September 2014, with each trip listed by date and pickup time, as well as GPS coordinates.

\begin{figure*}[htbp]
	\centering
	\foreach \sig in {603,2163,3057} {					
		\centering \subfloat{
			\centering \includegraphics[scale=0.4,clip,trim=1.65cm 0.8cm 4.7cm 0cm]{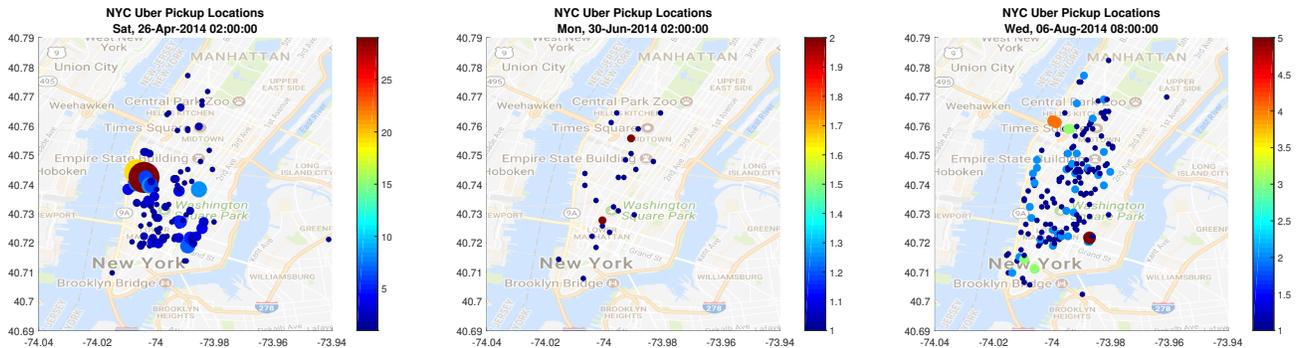}
		}
	}	
	\caption[]{Characteristic graph signals demonstrating the hourly number of Uber pickups at different locations in New York City. The size and color of each circle indicate the signal value at that graph node.}
	\label{Fig:UberSignals}
\end{figure*}

To create a graph from this raw data, we sampled the New York City region on a grid of $150\times 150$ points and assigned each pickup to its nearest grid point, accumulating the number of pickups in each grid location.
Similarly, pickups were aggregated over time intervals of one hour each, such that the total number of pickups in a specific hour is a graph signal. 
To enrich the graph structure, we selected only the subset of grid points for which the overall number of pickups exceeded 1000, keeping a total of $N=746$ nodes. 
The weight of the edge between the nodes $i$ and $j$ was set to $w_{ij}=\exp\left(\frac{-d^2(i,j)}{2\sigma^2}\right)$, 
where $d(i,j)$ is the Euclidean distance between their respective coordinates and $\sigma$ is a scaling factor. 
Exemplar signals of this dataset are illustrated in Figure~\ref{Fig:UberSignals}.

\begin{figure}[htb] 
	\centering
	\subfloat[]{
		\centering \includegraphics[scale=0.38,clip,trim=0.8cm 0.3cm 0.8cm 0.67cm]{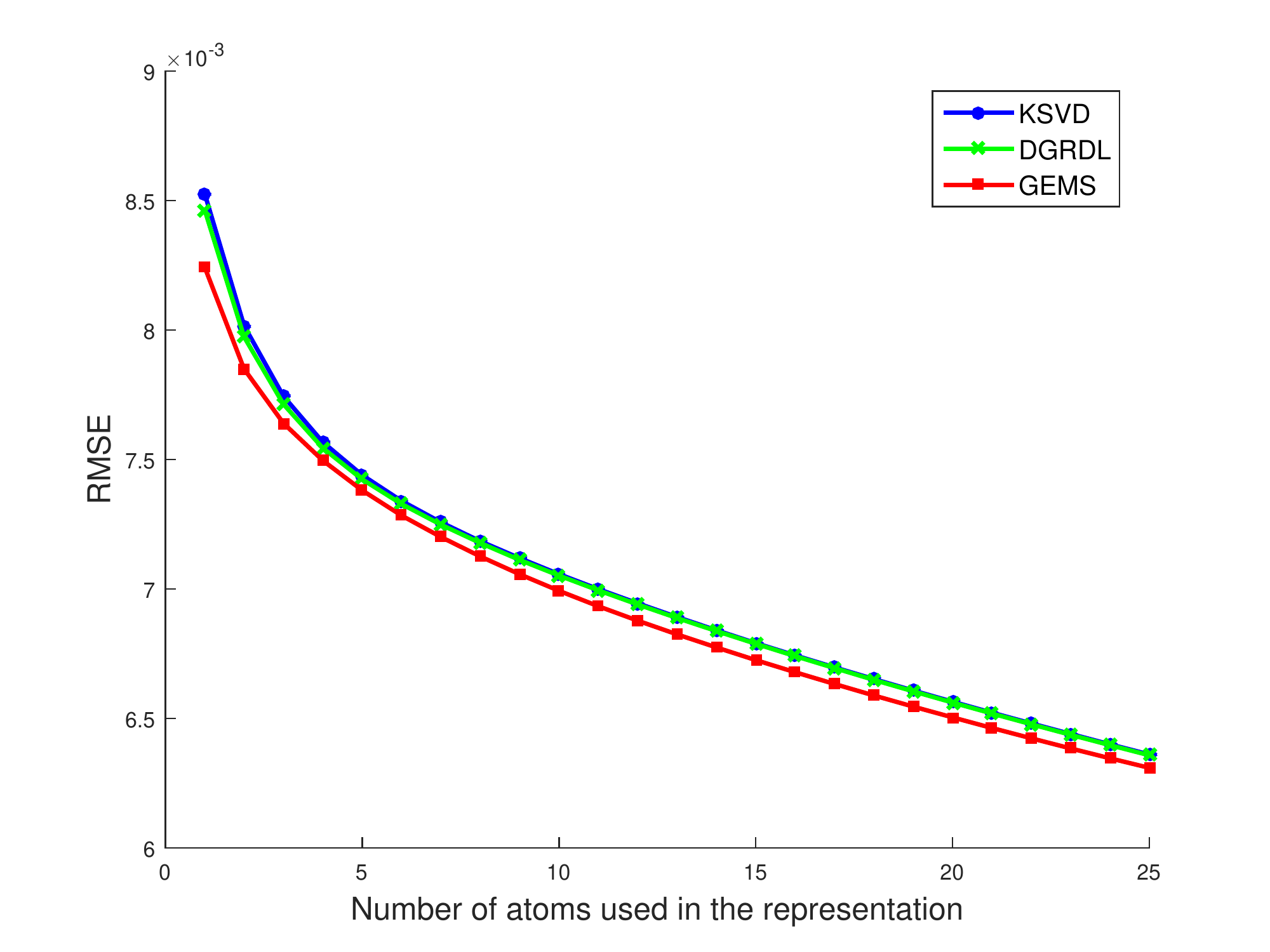}
		\label{Fig:uber_rep}
	}\\
	\subfloat[]{
		\centering \includegraphics[scale=0.38,clip,trim=0.8cm 0.3cm 0.8cm 0.67cm]{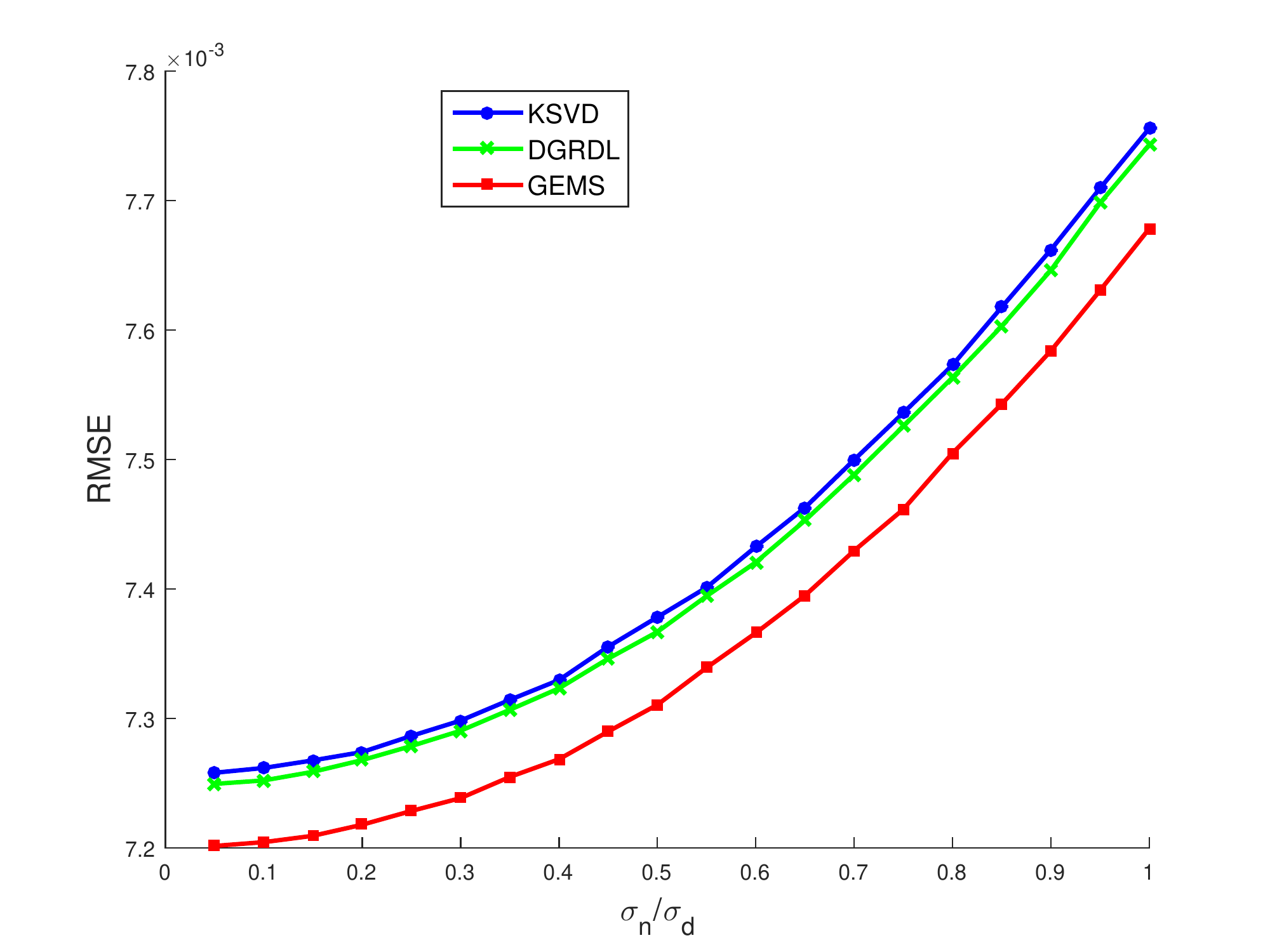}
		\label{Fig:uber_den}
	}
	\caption[]{Comparison of the learned dictionaries in terms of normalized RMSE for different applications tested on the Uber NYC pickups dataset: \subref{Fig:uber_rep} representation error for different sparsity levels, \subref{Fig:uber_den} denoising error for different noise levels $\sigma_n$ with respect to the data standard deviation $\sigma_d$.}
	\label{Fig:uber_rep_den}
\end{figure}

Following the previous experiments, we compared GEMS with K-SVD and DGRDL, which were the leading competitors. The different dictionaries were again trained and evaluated on the tasks of signal approximation and denoising.
Sparsity thresholds of $T=7$ and $P=30$ were used for training, and all signals were normalized with respect to the one having the maximal energy. 
The results are depicted in Figure~\ref{Fig:uber_rep_den}, establishing again the advantage of GEMS over the other compared methods.

\subsection{Discussion}
Just before we conclude this section, we would like to discuss an additional side benefit of the proposed GEMS algorithm. 
The multi-scale nature of the GEMS dictionary may serve data analysis tasks and be used for capturing important phenomena in the data. For instance, one might characterize and distinguish between different signals based solely on the dictionary atoms chosen for their approximation. Put differently, some of the learned patterns may be associated with a specific day of the week, or a specific time of day. 

\begin{figure*}[htb]
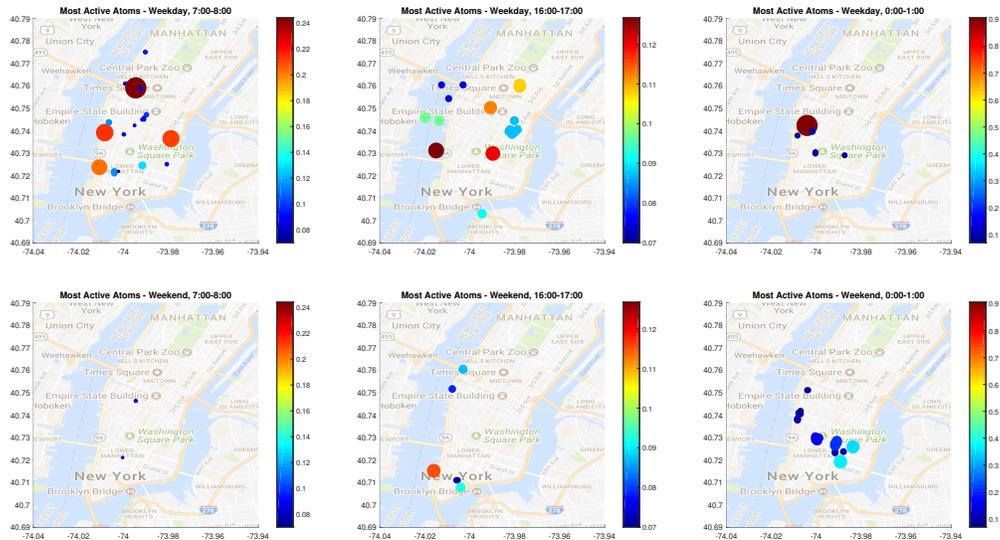
 
	\centering
	\foreach \hour in {7,16,0} {
		\centering \subfloat{
			\centering \includegraphics[scale=0.3,clip,trim=1.5cm 0.8cm 4.7cm 0cm]{Uber_weekday_\hour}
		}
	}\\
	\foreach \hour in {7,16,0} {
		\centering \subfloat{
			\centering \includegraphics[scale=0.3,clip,trim=1.5cm 0.8cm 4.7cm 0cm]{Uber_weekend_\hour}
		}
	}
	\caption[]{Comparing the most active GEMS atoms in representing Uber pickup counts on weekdays (top) and weekends (bottom), at different hours of the day (from left to right): 7-8AM, 4-5PM, 0-1AM.}
	\label{Fig:UberAnalysis}
\end{figure*}

To demonstrate this idea, we consider the signals measuring the number of Uber pickups during the times 7AM-8AM.
By sparse coding these signals over the trained dictionary and analyzing the chosen atoms statistics, we can distinguish between regions that are more active on weekdays and others that are more active on weekends. 
Repeating the experiment for other signal groups reveals the pattern variability between different hours of the day, as illustrated in Figure~\ref{Fig:UberAnalysis}.
\newline

As mentioned earlier, another essential property that the proposed dictionary structure introduces is locality and piecewise-smooth behavior. 
As advocated in \cite{Thanou2014}, for example, similar local patterns may appear at various locations across the network, and thus learning localized atoms may benefit the processing of high-dimensional graph signals. 

Indeed, graph signals emerging in various real-life applications are only piecewise-smooth (and not globally-smooth) over the graph.
For instance, while each community in a social network may have a relatively homogeneous behavior, some variability could be expected between communities, exhibiting delicate differences that the graph Laplacian cannot encode. 
Similarly, traffic patterns may be different in rural areas compared with urban regions and city centers, with sharper transitions occurring near city boundaries. 
Such phenomena are ill-represented by the graph Laplacian, even when inferred from the data. Since the Laplacian matrix models the common underlying structure of the given signals, it is often unable to account for the local nature of different network regions. 
In these cases, relying on a global smoothness is insufficient, and an alternative local (piecewise) regularity assumption may better fit such signals.

To highlight this property, the data in all the experiments presented above has a localized, clustered, or piecewise-smooth nature. 
As demonstrated throughout all the experiments, the global regularity assumption of DGRDL \cite{Yankelevsky2016} evidently makes it suboptimal for representing such signals. However, by relaxing this assumption and infusing a multi-scale structure to the learned dictionary, GEMS better applies to this broader class of graph signals.

\section{Conclusions} \label{Sec:Conclusions}
In this paper, we introduce a new dictionary learning algorithm for graph signals that mitigates the global regularity assumption and applies to a broader class of graph signals, while enabling treatment of higher dimensional data compared with previous methods.

The core concept of the proposed GEMS method lies in combining a simple and efficient graph-Haar wavelet basis, that brings a multi-scale nature we deem vital for representing large signals, with a learned sparse component, that makes it adaptive to the given data.

The underlying graph topology is incorporated in two manners. 
The first is implicit, by modeling the learned dictionary atoms as sparse combinations of graph wavelet functions, thus practically designing an adaptable multi-scale dictionary. 
The second is explicit, by adding direct graph constraints to preserve the local geometry and promote smoothness in both the feature and manifold domains.  

Furthermore, the complete optimization scheme offers the ability to refine the graph Laplacian $L$, as well as the graph-wavelet basis $\Phi$, as an integral part of the dictionary learning process.

The effectiveness of the proposed algorithm is demonstrated through experiments on both synthetic data and real network data, showing that it achieves superior performance to other tested methods in data processing and analysis applications of different nature and different dimensions.

\section*{Acknowledgments}
The research leading to these results has received funding from the European Research Council under European Union's Seventh Framework Program, ERC Grant agreement no. 320649, and from the Israel Science Foundation (ISF) grant number 1770/14.

\bibliographystyle{IEEEtran}
\bibliography{GEMS_bibfile}
\end{document}